\RequirePackage{fix-cm}
\documentclass[smallextended]{svjour3}       % onecolumn (second format)
\smartqed  % flush right qed marks, e.g. at end of proof
\usepackage{amsmath}
\usepackage{amsfonts}
\usepackage{wrapfig}
\usepackage{hyperref}

\usepackage{color}
\usepackage{graphicx}
\usepackage[paperwidth=7in, paperheight=10in, margin=.875in]{geometry}
\begin{document}

\title{Nonlinear Dynamical Systems for Automatic Face Annotation in Head Tracking and Pose Estimation}

%\subtitle{Nonlinear Dynamical Systems}

\titlerunning{Nonlinear Dynamical Systems}        % if too long for running head

\author{Thoa Thieu        \and
        Roderick Melnik 
}

%\authorrunning{Short form of author list} % if too long for running head

\institute{Thoa Thieu \at
              School of Mathematical and Statistical Sciences, The University of Texas Rio Grande Valley, \\ Edinburg, Texas, USA
              \\
               \email{thoa.thieu@utrgv.edu}           %  \\
%             \emph{Present address:} of F. Author  %  if needed
           \and
           Roderick Melnik \at
              MS2Discovery Interdisciplinary Research Institute, Wilfrid Laurier University, \\ Waterloo, Ontario, Canada \\ \email{rmelnik@wlu.ca}
}

\date{Received: date / Accepted: date}
% The correct dates will be entered by the editor

\maketitle

\begin{abstract}
Facial landmark tracking plays a vital role in applications such as facial recognition, expression analysis, and medical diagnostics. In this paper, we consider the performance of the Extended Kalman Filter (EKF) and Unscented Kalman Filter (UKF) in tracking 3D facial motion in both deterministic and stochastic settings. We first analyze a noise-free environment where the state transition is purely deterministic, demonstrating that UKF outperforms EKF by achieving lower mean squared error (MSE) due to its ability to capture higher-order nonlinearities. However, when stochastic noise is introduced, EKF exhibits superior robustness, maintaining lower mean square error (MSE) compared to UKF, which becomes more sensitive to measurement noise and occlusions. Our results highlight that UKF is preferable for high-precision applications in controlled environments, whereas EKF is better suited for real-world scenarios with unpredictable noise. These findings provide practical insights for selecting the appropriate filtering technique in 3D facial tracking applications, such as motion capture and facial recognition.

% \PACS{PACS code1 \and PACS code2 \and more}
%\subclass{MSC code1 \and MSC code2 \and more}
\end{abstract}

\section{Introduction}
\label{intro}
Accurate 3D facial tracking has become an essential tool in fields such as computer vision, human-computer interaction, and biometric security. The demand for real-time, precise tracking of facial movements has increased with the growing need for applications like facial expression recognition, virtual avatars, emotion analysis, and advanced video conferencing systems. However, achieving high-quality 3D facial tracking is inherently challenging due to the complexity of human facial dynamics, which involve numerous nonrigid deformations and various lighting and occlusion conditions. As such, researchers have turned to advanced filtering techniques like Kalman Filters (KFs) to estimate the dynamic state of facial models with reduced error and uncertainty.

The KF is a highly effective algorithm for estimating and predicting the states of a system under both deterministic and stochastic uncertainty. It has been widely applied across various fields, including target tracking, navigation, and control systems, where accurate state estimation is crucial despite the presence of noise and unpredictable factors \cite{Khodarahmi2023,Yadav2023,Wu2002neural}. As the optimal linear estimator for systems described by linear models, the Kalman filter assumes additive independent white noise in both process and measurement dynamics. However, many real-world applications—ranging from biological systems to engineering—are inherently nonlinear, requiring extensions of the standard Kalman filter to handle these complexities.

To address the challenges of nonlinear state estimation, researchers have developed advanced filtering techniques such as the Extended Kalman Filter (EKF) and the Unscented Kalman Filter (UKF). EKF, which relies on linearization techniques, has proven effective for systems with mild nonlinearity, whereas UKF employs sigma point propagation, making it better suited for more complex nonlinearities \cite{Wan2000unscented,St2004comparison,Singh2023inverse}. Nevertheless, the performance of these filters remains sensitive to noise characteristics, including sensor errors, occlusions, and sudden movements, which frequently occur in real-world scenarios \cite{Law2015data,Toivanen2016,Yang2023outlier,Pyrhonen2023}.

These filtering techniques have been extensively applied in human pose estimation and facial tracking tasks. Agarwal et al. \cite{Agarwal2013estimating} employed UKF for human pose estimation from monocular video, refining lower-limb dynamics by filtering out inaccurate pose estimates in dynamic environments. Ardiyanto et al. \cite{Ardiyanto2014partial} integrated partial least squares models with UKF-based tracking to estimate upper body orientation, demonstrating the effectiveness of hybrid filtering approaches. In facial tracking, Yang et al. \cite{Yang2016cascaded} utilized Kalman filtering in their cascaded elastically progressive model (EPM) for face alignment, which significantly improved landmark localization in challenging conditions such as occlusions and varying lighting. Similarly, Dalimarta et al. \cite{Dalimarta2021lower} developed a lower-body detection and tracking framework using AlphaPose and EKF, achieving high accuracy (95.17\%) in detecting lower-leg movements. Beyond human applications, Kuncara et al. \cite{Kuncara2024integration} introduced the nonlinear observer-unscented Kalman filter (NLO-UKF) to enhance state estimation accuracy in highly complex, nonlinear systems such as autonomous trucks. Ramadan et al. \cite{Ramadan2024} highlighted that while the EKF provides an accurate Bayesian framework for certain problems, alternative filters may be needed for systems with more complex uncertainties.

Head pose estimation, a critical component of computer vision, has also benefited from these advancements. It plays a crucial role in applications such as human-computer interaction, surveillance, and autonomous driving. Comprehensive surveys by Khan et al. \cite{Khan2021head} and Abate et al. \cite{Abate2022head} provide insights into the evolution of head pose estimation, covering both traditional and deep learning-based approaches. Among recent innovations, Hu et al. \cite{Hu2022toward} proposed a spatiotemporal vision transformer for dynamic head tracking, leveraging deep learning to improve accuracy. Fang et al. \cite{Fang2022alphapose} introduced AlphaPose, a real-time whole-body multi-person pose estimation system that enhances tracking through techniques such as Symmetric Integral Keypoint Regression and Pose-Aware Identity Embedding. Additionally, Joska et al. \cite{Joska2021acinoset} demonstrated the versatility of pose estimation methods beyond human applications with AcinoSet, a dataset designed for 3D pose tracking of cheetahs in the wild. These contributions collectively advance the field by refining pose estimation techniques, improving robustness, and expanding their applicability to diverse domains.

Pose estimation is not only relevant to human-centered applications but also plays a vital role in spaceborne vision-based navigation, particularly when tracking non-cooperative targets. Wang et al. \cite{Wang2024monocular} proposed a monocular satellite pose estimation method that integrates uncertainty modeling and self-assessment to enhance accuracy, efficiency, and reliability in real-time applications. Their approach employs a lightweight neural network for keypoint prediction and refines pose estimation through aleatoric uncertainty modeling and entropy-based self-assessment. Similarly, Park et al. \cite{Park2024online} addressed the domain gap challenge in neural network-based spaceborne navigation by introducing an Online Supervised Training (OST) framework. Their method leverages an adaptive UKF to improve neural network performance in-flight, thereby enhancing pose estimation accuracy during rendezvous and proximity operations. These advancements contribute to more robust and reliable spacecraft pose estimation, which is essential for autonomous space missions.

From human tracking to spaceborne applications, the evolution of Kalman filtering techniques and deep learning-based pose estimation methods continues to drive innovation in state estimation. By integrating uncertainty modeling, adaptive filtering, and neural networks, researchers are developing more accurate and resilient tracking systems that can operate effectively in highly dynamic and unpredictable environments.

This paper aims to provide a detailed comparison of EKF and UKF in the context of 3D facial tracking, addressing both deterministic and stochastic environments. We evaluate the performance of the EKF and the UKF in reconstructing the temporal evolution of 54 facial landmarks. The real landmark positions are obtained from the dataset presented in \cite{Ariz2016novel}, consisting of sequential 3D face models. The estimated landmark coordinates are computed using state-space modeling, and the results illustrate the accuracy of both filtering techniques in tracking facial landmarks over time. The plotted facial landmark coordinates highlight how the estimation aligns with real data, providing insights into the effectiveness of each method.  We explore the advantages and limitations of each filter in terms of tracking accuracy, noise resilience, and computational efficiency. By examining both noise-free and noisy scenarios, we highlight the factors that influence the performance of these algorithms and offer insights into their practical applications. Furthermore, we propose potential strategies to enhance the robustness of UKF in challenging conditions, such as hybrid filtering approaches and parameter tuning.
% This research contributes to the advancement of precise and reliable facial motion tracking and lays the groundwork for future investigations in stochastic scenarios.
The results presented in this work are intended to assist researchers and practitioners in selecting the appropriate Kalman filtering technique for their 3D facial tracking systems, depending on the specific requirements of their application.

	\section{Model description}\label{model-description}
	In this section, we present two widely used state estimation techniques: the EKF and UKF. These methods are designed to estimate the state of a dynamic system from noisy observations while accounting for uncertainty in the system's evolution and measurements. The EKF approximates nonlinear dynamics by linearizing the system at each step using Jacobian matrices, whereas the UKF employs the unscented transform to propagate uncertainty more accurately without relying on linearization. We provide a detailed description of both models, outlining their mathematical formulation and key algorithmic steps \cite{Law2015data,Wan2000unscented}.
	In this work, we extend the application of these filtering techniques to the challenging task of 3D facial landmark tracking, where precise estimation of dynamic facial movements is crucial. A key novelty of our approach is the systematic evaluation of EKF and UKF performance across both deterministic and stochastic scenarios, analyzing their strengths and weaknesses in reconstructing the temporal evolution of 54 facial landmarks obtained from the dataset in \cite{Ariz2016novel}.
	
	These improvements are particularly relevant in real-world facial motion tracking applications, where filtering accuracy directly impacts downstream tasks such as expression recognition and 3D face reconstruction. The numerical results in Section \ref{numerical-results} illustrate how these modifications improve the reliability of facial landmark estimation, providing insights into the practical deployment of these filters.
	\subsection{EKF model}
In this section, we introduce a model of stochastic dynamics in the EKF. In the
context of the EKF, the system’s evolution and the measurement process are
both inherently stochastic, meaning they involve random variables or uncertainties. The EKF is designed to handle nonlinear stochastic systems by predicting the system's state and estimating the uncertainty in the state over time. It updates its state estimate based on noisy observations, effectively modeling how uncertainty evolves as the system progresses.

The EKF algorithm is recursive, meaning it updates the state estimate at each time step based on previous estimates and new observations. The key feature of the EKF, as opposed to the traditional Kalman Filter (KF), is that it can handle nonlinear models by linearizing them at each time step around the current estimate of the state. Below are the key steps of the EKF prediction and update processes (see, e.g. \cite{Law2015data,Wan2000unscented}).

\subsubsection{State prediction:} 
	
%	\[
%	\mathbf{x}_{k|k-1} = \mathbf{x}_{k-1|k-1} + f(\mathbf{x}_{k-1|k-1}) \cdot \Delta t + \mathbf{w}_k,
%	\]

\[
\mathbf{x}_{k+1} = f(\mathbf{x}_k, \mathbf{w}_k, \Delta t),
\]
	where \(\mathbf{x}_{k|k-1}\) is the predicted state at time step \(k\),$f(\mathbf{x}_{k-1|k-1})$  denotes the nonlinear process model,  \(\mathbf{w}_k \sim \mathcal{N}(0, Q)\) stands for process noise (with covariance \(Q\)), while  \(\Delta t\) represents the time step between two successive measurements or updates.
	In the state prediction step, we predict the state of the system at the current time step $k$ (\(\mathbf{x}_{k|k-1}\))based on the previous state estimate $\mathbf{x}_{k-1|k-1}$. The term $f(\mathbf{x}_{k-1|k-1})$ describes how the system evolves over time. It is applied to the previous state to estimate the system's state at time $k$.
\subsubsection{Covariance prediction:}
	\[
	P_{k|k-1} = F_{k-1} \cdot P_{k-1|k-1} \cdot F_{k-1}^T + Q,
	\]
	where \(P_{k|k-1}\) is the predicted covariance, \(F_{k-1}\) denotes the Jacobian of the process model at the previous state, i.e., \(F_{k-1} = \frac{\partial f(\mathbf{x})}{\partial \mathbf{x}}\),
	and \(P_{k-1|k-1}\) represents the covariance matrix from the previous time step.
	
\subsubsection{Measurement update:}  
	
%	\[
%	\mathbf{z}_{k} = h(\mathbf{x}_{k|k-1}) + \mathbf{v}_k,
%	\]
	\[
\mathbf{z}_{k} = h(\mathbf{x}_k,\mathbf{v}_k),
\]
	where \(h(\cdot)\) is the nonlinear measurement function, and \(\mathbf{v}_k \sim \mathcal{N}(0, R)\) is measurement noise (with covariance \(R\)).
	\subsubsection{ Kalman gain:}
	\[
	K_k = P_{k|k-1} \cdot H_k^T \cdot (H_k \cdot P_{k|k-1} \cdot H_k^T + R)^{-1},
	\]
	where \(H_k\) is the Jacobian of the measurement model, i.e., \(H_k = \frac{\partial h(\mathbf{x})}{\partial \mathbf{x}}\). The Kalman gain $K_k$ determines how much the state estimate should be corrected based on the new measurement. It is computed by using the predicted covariance $P_{k|k-1}$ and the Jacobian $H_k^T$ of the measurement model. The Kalman gain is designed to minimize the estimation error by balancing the uncertainty in the predicted state and the measurement. A high Kalman gain indicates that the measurement is trusted more, while a low gain indicates that the predicted state is more reliable.
	
\subsubsection{ State update:}
	\[
	\mathbf{x}_{k|k} = \mathbf{x}_{k|k-1} + K_k \cdot (\mathbf{z}_k - h(\mathbf{x}_{k|k-1})),
	\]
	where \(\mathbf{x}_{k|k}\) is the updated state estimate at time \(k\),
	and \(\mathbf{z}_k\) denotes the actual measurement at time \(k\). 
	
	\subsubsection{ Covariance update:}
	\[
	P_{k|k} = (I - K_k \cdot H_k) \cdot P_{k|k-1},
	\]
	where \(I\) is the identity matrix of appropriate size. This step ensures that the covariance matrix reflects the new state estimate's uncertainty after the measurement correction.

The EKF is an iterative algorithm that recursively updates the state estimate and its uncertainty. It does so by predicting the state based on the process model, updating the prediction based on noisy measurements, and using the Kalman gain to combine these two sources of information. The nonlinearity of the system and measurement models is handled by linearizing the models at each time step using Jacobian matrices. 

\subsection{UKF model}

Building on the principles of the EKF, the UKF presents an improved approach for dealing with nonlinear systems. Unlike the EKF, which uses linearization via Jacobian matrices to approximate the system and measurement models, the UKF propagates uncertainty by using a set of carefully selected sample points called sigma points. These points are chosen to better capture the nonlinearities inherent in the system and measurement models, resulting in more accurate estimates. In the following section, we provide a detailed description of the UKF, explaining how it overcomes the limitations of the EKF and offers a more robust method for nonlinear state estimation. The UKF follows two key steps: the prediction step and the update step, which iteratively estimate the state using measurements (see, e.g. \cite{Law2015data,Wan2000unscented}).

\subsubsection{State evolution (process model)}

The process model describes how the state evolves over time. Let \( \mathbf{x}_k \) represent the state at time step \( k \), and it is assumed to evolve according to the following equation:
%	\[
%\mathbf{x}_{k+1|k} = \mathbf{x}_{k} + f(\mathbf{x}_{k}) \cdot \Delta t + \mathbf{w}_k,
%\]
\[
\mathbf{x}_{k+1} = f(\mathbf{x}_k, \mathbf{w}_k, \Delta t),
\]
where \( \mathbf{x}_k \) is the state vector at time \( k \),  \( f(\cdot) \) denotes the nonlinear state transition function (process model),  \( \mathbf{w}_k \) is the process noise (assumed to be Gaussian with zero mean and covariance \( Q \)), and \( \Delta t \) is the time step between measurements.

\subsubsection{Measurement model}

The measurement model relates the true state to the observed measurements. It is typically expressed as:

\[
\mathbf{z}_k = h(\mathbf{x}_k, \mathbf{v}_k),
\] where \( \mathbf{z}_k \) is the measurement vector at time \( k \),  \( h(\cdot) \) is the nonlinear measurement function, and \( \mathbf{v}_k \) is the measurement noise (assumed to be Gaussian with zero mean and covariance \( R \)).

\subsubsection{Unscented transform (sigma points generation)}

The UKF propagates the uncertainty in the state estimate by using sigma points. These are carefully selected points that represent the distribution of the state vector. The idea is to generate a set of points around the current state estimate that captures the state’s uncertainty more accurately than a simple linear approximation.

Given the current state estimate \( \hat{\mathbf{x}}_k \) and the covariance \( P_k \), the sigma points \( \mathbf{\chi}_k \) are calculated using the following formulas:

\[
\mathbf{\chi}_0 = \hat{\mathbf{x}}_k
\]

\[
\mathbf{\chi}_i = \hat{\mathbf{x}}_k + \left( \sqrt{(n + \lambda) P_k} \right)_i, \quad i = 1, \ldots, n
\]

\[
\mathbf{\chi}_{i+n} = \hat{\mathbf{x}}_k - \left( \sqrt{(n + \lambda) P_k} \right)_i, \quad i = 1, \ldots, n,
\]
where
 \( n \) is the dimension of the state, and \( \lambda \) is a scaling parameter (used to control the spread of the sigma points).

\subsubsection{Prediction step}

Once the sigma points are generated, they are propagated through the process model to obtain the predicted sigma points. Each sigma point \( \mathbf{\chi}_i \) is propagated through the process model:

\[
\mathbf{\chi}_{k|k+1}^{(i)} = f(\mathbf{\chi}_{k-1}^{(i)}, \Delta t),
\]
where \( f(\cdot) \) is the process model function. The predicted state \( \hat{\mathbf{x}}_{k+1|k} \) is then computed as the weighted mean of the predicted sigma points:

\[
\hat{\mathbf{x}}_{k|k-1} = \sum_{i=0}^{2n} W^{(m)}_i \mathbf{\chi}_{k|k-1}^{(i)},
\]
where \( W^{(m)}_i \) are the weights associated with the mean. Here, the weights for the mean are:
\[
W^{(m)}_0 = \frac{\lambda}{n + \lambda}, \quad W^{(m)}_i = \frac{1}{2(n + \lambda)}, \quad i = 1, \ldots, 2n.
\]

The predicted covariance \( P_{k|k-1} \) is updated as:

\[
P_{k|k-1} = \sum_{i=0}^{2n} W^{(c)}_i (\mathbf{\chi}_{k|k-1}^{(i)} - \hat{\mathbf{x}}_{k|k-1}) (\mathbf{\chi}_{k|k-1}^{(i)} - \hat{\mathbf{x}}_{k|k-1})^T + Q,
\]
where \( W^{(c)}_i \) are the weights associated with the covariance, while \( Q \) is the process noise covariance matrix. The predicted covariance is calculated by taking the weighted sum of the outer products of the differences between the predicted sigma points and the predicted state.

\subsubsection{Update step}

When a new measurement \( \mathbf{z}_k \) is received, the goal is to update the state estimate \( \hat{\mathbf{x}}_{k|k-1} \) and its covariance \( P_{k|k-1} \).

\begin{itemize}
	\item[a. ] Predicted measurement: The predicted measurement is calculated by applying the measurement model \( h(\cdot) \) to the predicted sigma points:
	
	\[
	\mathbf{z}_{k|k-1}^{(i)} = h(\mathbf{\chi}_{k|k-1}^{(i)}).
	\] 
	\item[b. ] Measurement prediction: The predicted measurement \( \hat{\mathbf{z}}_{k+1|k} \) is the weighted mean of the predicted measurements:
	
	\[
	\hat{\mathbf{z}}_{k|k-1} = \sum_{i=0}^{2n} W^{(m)}_i \mathbf{z}_{k|k-1}^{(i)}.
	\]
	\item[c. ]  Innovation: The innovation is the difference between the actual measurement \( \mathbf{z}_k \) and the predicted measurement \( \hat{\mathbf{z}}_{k|k-1} \):
	
	\[
	\mathbf{y}_k = \mathbf{z}_k - \hat{\mathbf{z}}_{k|k-1}.
	\] 
	\item[d. ]  Innovation covariance: The covariance of the innovation is computed as:
	
	\[
	S_k = \sum_{i=0}^{2n} W^{(c)}_i (\mathbf{z}_{k|k-1}^{(i)} - \hat{\mathbf{z}}_{k|k-1}) (\mathbf{z}_{k|k-1}^{(i)} - \hat{\mathbf{z}}_{k|k-1})^T + R,
	\]
	where \( R \) is the measurement noise covariance matrix. The innovation covariance represents the uncertainty in the measurement prediction, and it is used to adjust the Kalman gain.
	\item[e. ] Cross-covariance: The cross-covariance between the state and the measurement is:
	
	\[
	P_{xz} = \sum_{i=0}^{2n} W^{(c)}_i (\mathbf{\chi}_{k|k-1}^{(i)} - \hat{\mathbf{x}}_{k|k-1}) (\mathbf{z}_{k|k-1}^{(i)} - \hat{\mathbf{z}}_{k|k-1})^T. 
	\] The cross-covariance measures how the state and measurement co-vary, providing information about their relationship.
	\item[f. ]  Kalman gain: The Kalman gain \( K_k \) is computed as:
	
	\[
	K_k = P_{xz} S_k^{-1}.
	\] The Kalman gain determines how much the predicted state should be corrected based on the measurement innovation.
	\item[g. ] State update: The updated state estimate \( \hat{\mathbf{x}}_{k+1|k+1} \) is:
	
	\[
	\hat{\mathbf{x}}_{k|k} = \hat{\mathbf{x}}_{k|k-1} + K_k \mathbf{y}_k.
	\] The state estimate is updated by adding the innovation weighted by the Kalman gain.
	\item[h. ] Covariance update: The updated covariance \( P_{k|k} \) is:
	
	\[
	P_{k|k} = P_{k|k-1} - K_k S_k K_k^T.
	\] The state covariance is updated to reflect the reduced uncertainty after incorporating the new measurement.
	
\end{itemize}

%These steps collectively allow the UKF to efficiently estimate the state of a nonlinear system, even in the presence of process and measurement noise. The UKF's ability to propagate uncertainty through sigma points makes it more accurate and robust compared to the EKF, especially in highly nonlinear systems.

\section{Mean Squared Error (MSE)} 

The MSE measures the average of the squared differences between the predicted and true values. For each time step \( k \), the MSE is computed as:

\[
\text{MSE}_k = \frac{1}{N} \sum_{i=1}^{N} \left( \hat{\mathbf{x}}_{k,i} - \mathbf{x}_{k,i} \right)^2,
\] where  \( N \) is the number of state dimensions (again, \( N = 3 \times \text{num points} \)),  \( \hat{\mathbf{x}}_{k,i} \) denotes the estimated value of the \( i \)-th point at time step \( k \), while \( \mathbf{x}_{k,i} \) is the true value (real data) of the \( i \)-th point at time step \( k \).

The MSE gives a measure of the variance of the prediction error, and larger errors have a disproportionate effect due to the squaring of differences. MSE is the average of the squared differences between the predicted and true values, which is more sensitive to large errors than MAE and gives an indication of the error variance.

%	- MAE is the average of the absolute differences between the predicted and true values, and it provides a straightforward measure of error in the same units as the original data.

%The mathematical framework involves state prediction, measurement update, and the use of the Unscented Kalman Filter (UKF) to estimate the 3D positions of facial landmarks. The modified UKF includes a model with acceleration dynamics, offering a more flexible and accurate tracking model. By running multiple realizations, the algorithm averages out noise and provides a more stable estimate. The accuracy of the estimates is evaluated using error metrics like MAE and RMSE.

Both algorithms provide a framework for estimating the state of a nonlinear dynamic system based on noisy observations. The EKF approximates the nonlinearity by linearizing the process and measurement models, while the UKF avoids linearization by using sigma points to capture the mean and covariance of the state.

\section{Numerical results}\label{numerical-results}

\begin{figure}[h!]
	\centering
	\begin{tabular}{cc}
		\includegraphics[width=0.45\textwidth, height=5cm]{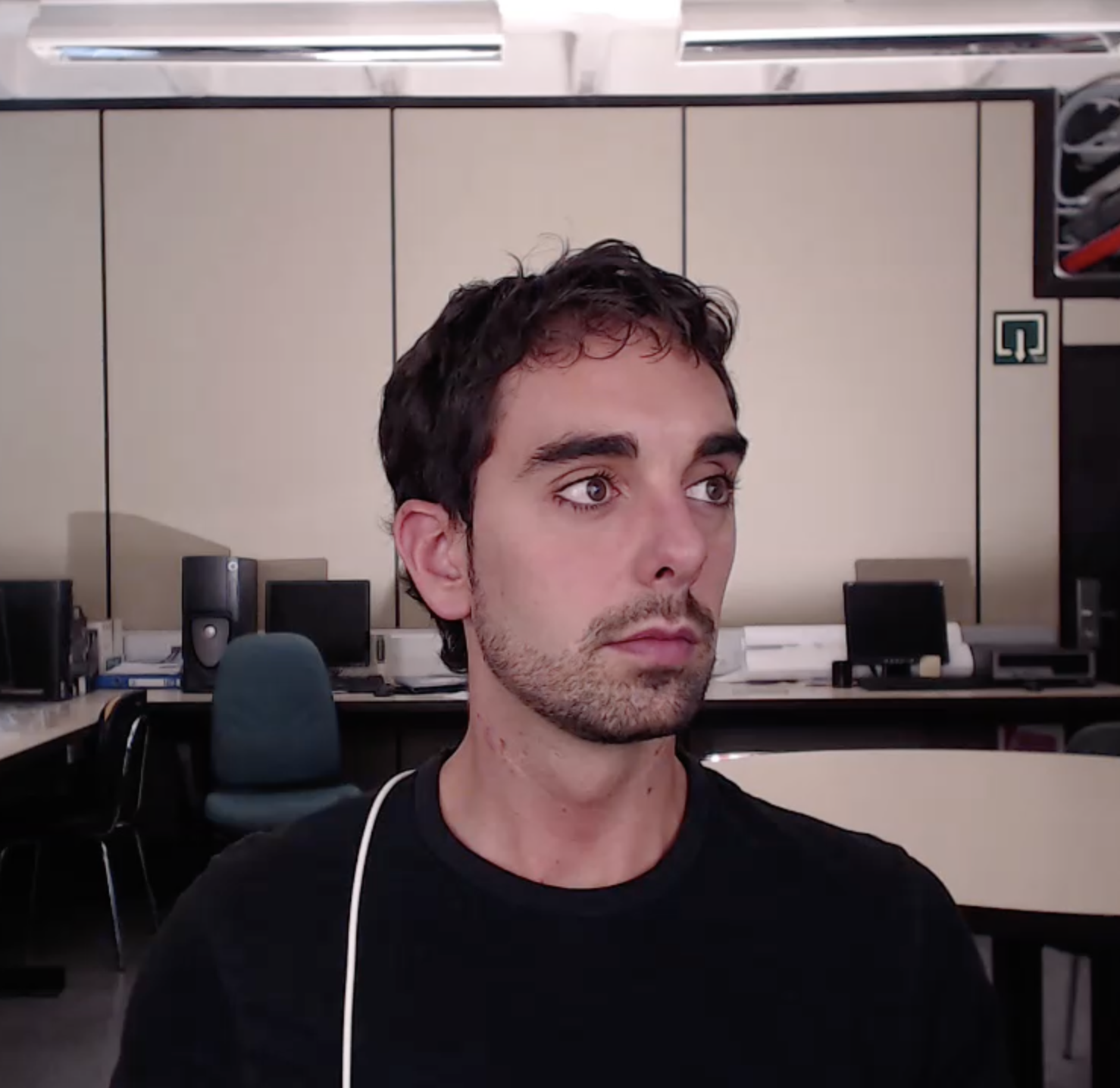} &
		\includegraphics[width=0.45\textwidth, height=5cm]{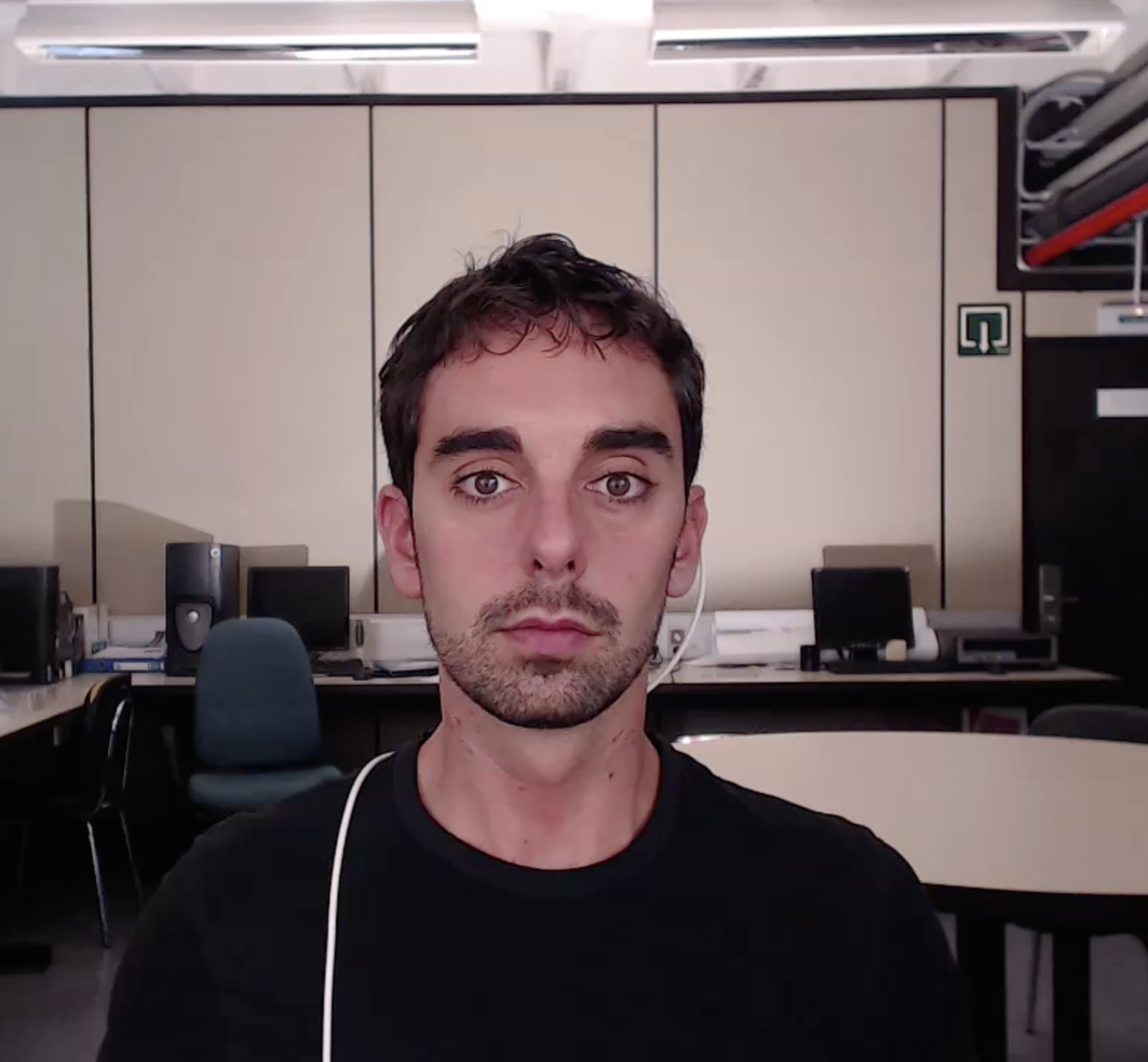} \\
		\includegraphics[width=0.45\textwidth, height=5cm]{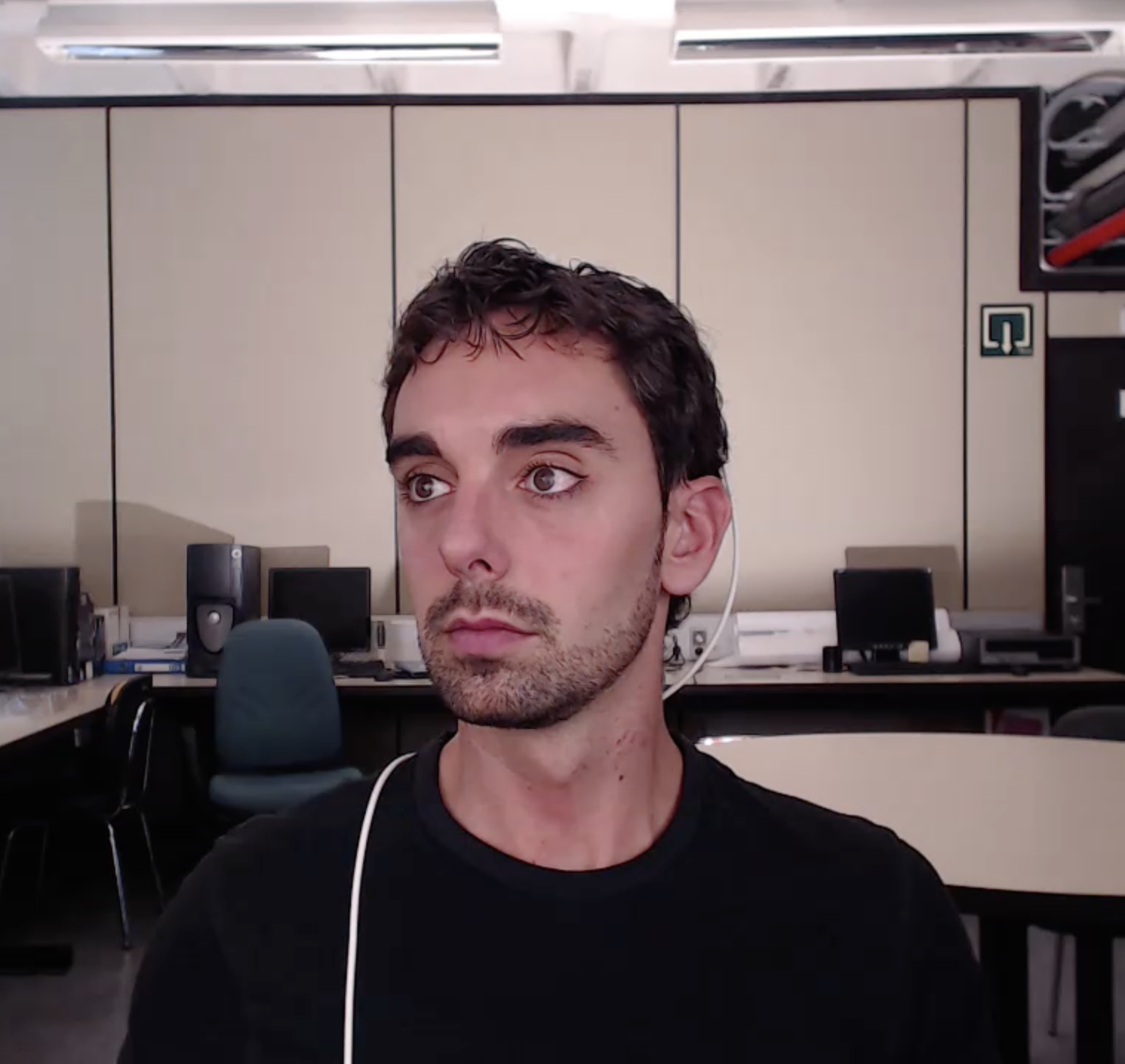} &
		\includegraphics[width=0.45\textwidth, height=5cm]{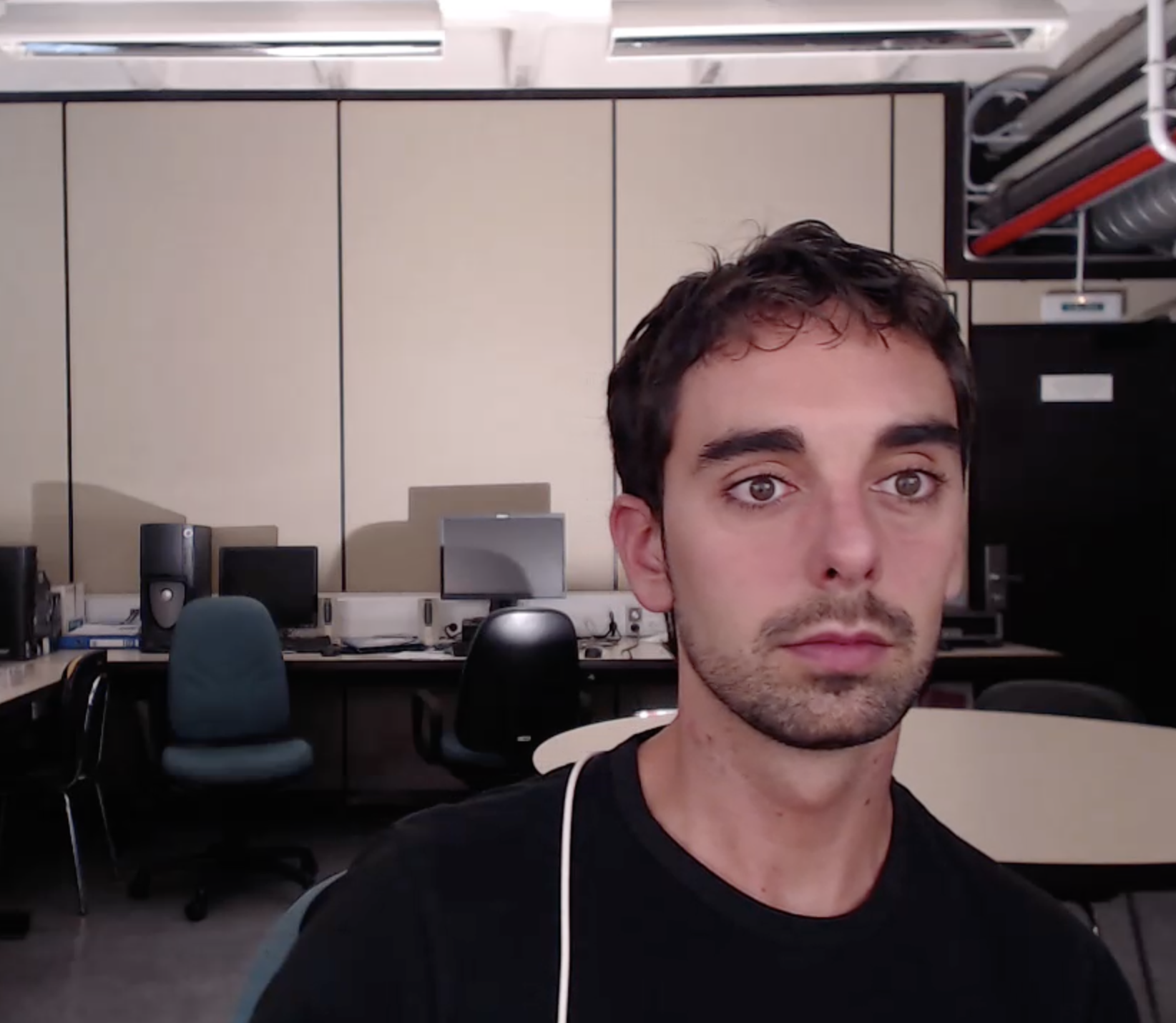} \\
		\includegraphics[width=0.45\textwidth, height=5cm]{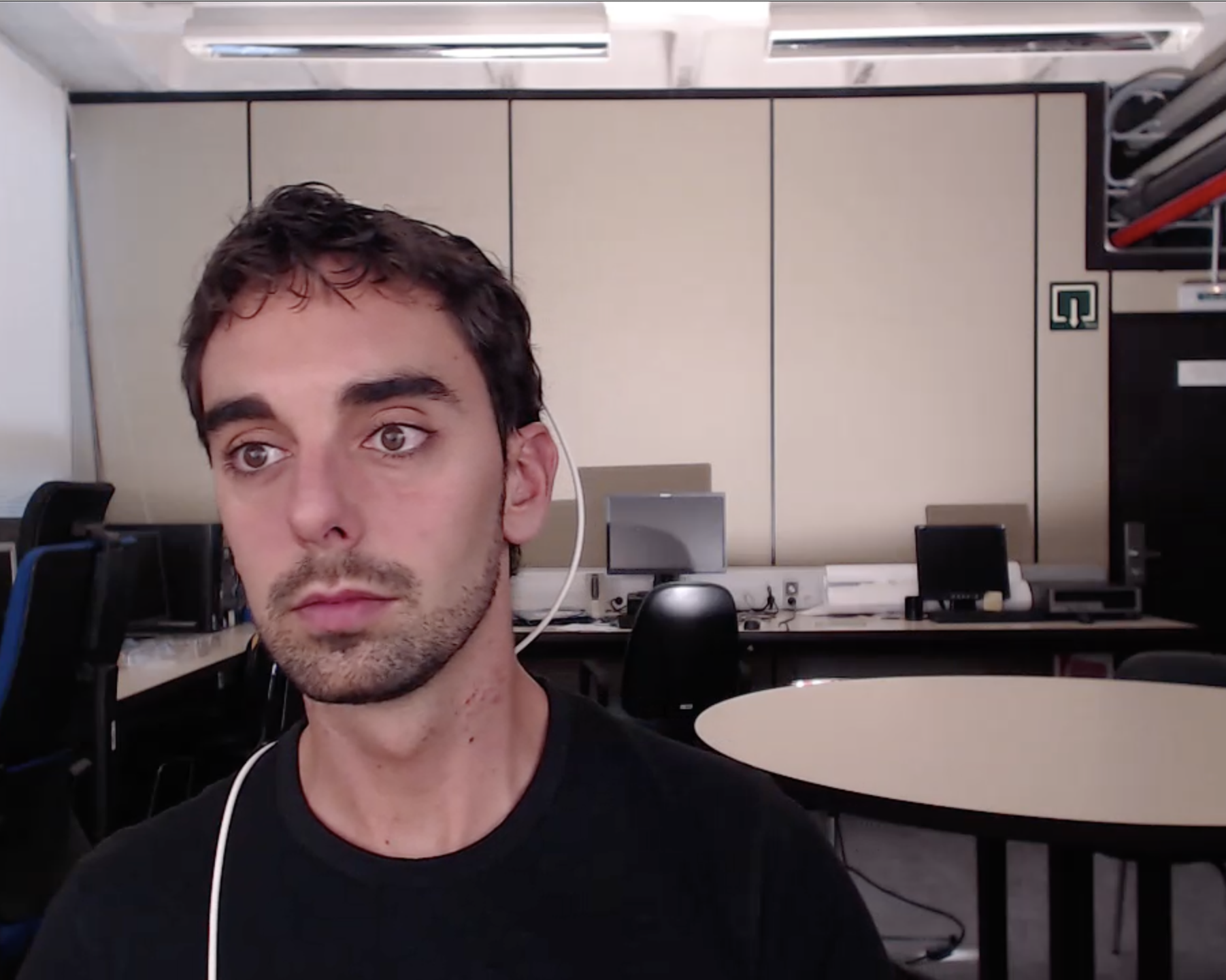} &
		\includegraphics[width=0.45\textwidth, height=5cm]{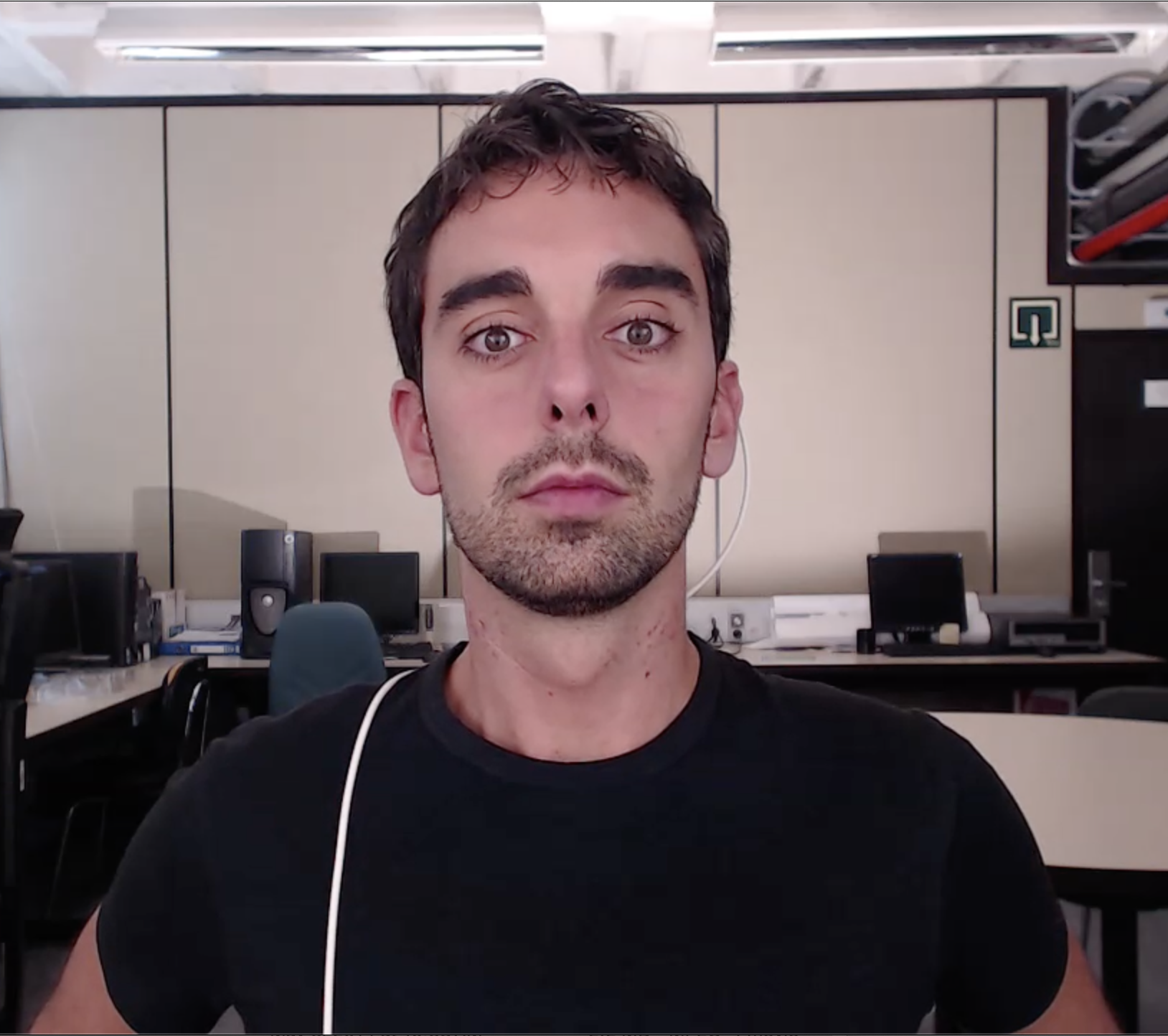}
	\end{tabular}
	\caption{[Color online] Illustration of the database for the first user. Top left panel: The man moves his face to the right. Top right panel: The man returns to the initial position, moving from the right. Middle left panel: The man moves his face to the left. Middle right panel: The man holds his face vertically and moves horizontally to the right. Bottom left panel: The man holds his face vertically and moves horizontally to the left. Bottom right panel: The man moves his face upward. The screenshots used in this illustration are provided in videos from the dataset in \cite{Ariz2016novel}, licensed under a Creative Commons Attribution-NonCommercial-ShareAlike 4.0 International License.} 
	\label{fig:0}
\end{figure}

\begin{figure}[h!]
	\centering
	\includegraphics[width=0.65\textwidth]{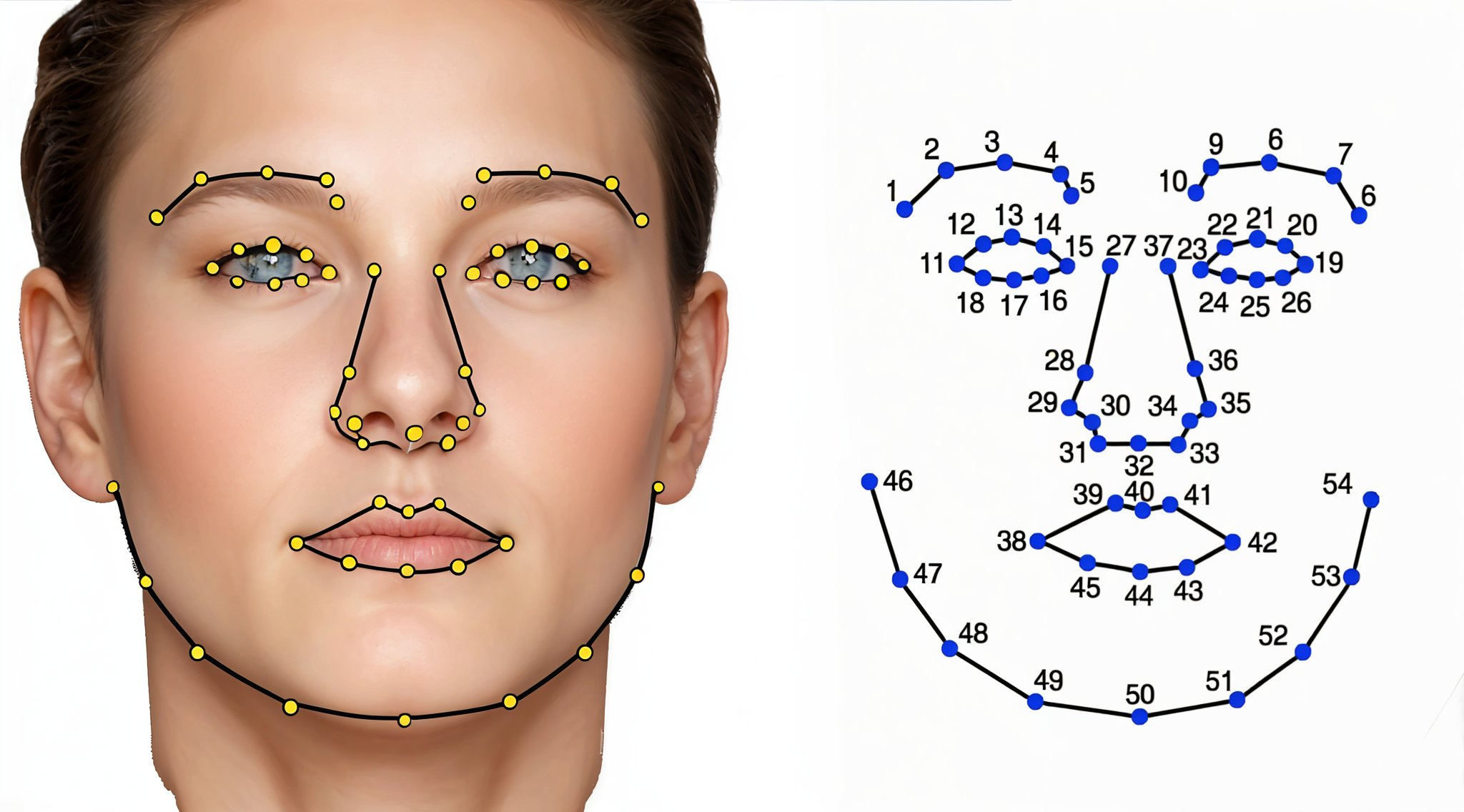} 
	\caption{[Color online] Illustration of the 54 facial landmarks automatically annotated for the dataset provided in \cite{Ariz2016novel}, showing both the anatomical placement (left) and numerical order (right) of the landmarks. This figure was created using Leonardo AI, Ibis Paint X, and Freepik Retouch.}
	\label{fig:1}
\end{figure}

In this section, we consider two case studies on the application of EKF and UKF, as described in Section \ref{model-description}, in a 3D facial tracking scenario. The dataset used in this study consists of 120 videos featuring 10 users (6 males and 4 females), with each subject contributing 12 recordings \cite{Ariz2016novel}. The real 3D face model files contain 54 rows, each corresponding to one 3D facial point, and 3 columns, corresponding to the $x, y$ and $z$ coordinates. Coordinate values are in millimeters. The points are referenced to the same sensor that acquires the head pose (see, e.g. Figs \ref{fig:0}-\ref{fig:1}). All simulations were conducted using Python, and the key numerical results indicate that EKF effectively mitigates noise, leading to a significant improvement in tracking accuracy across both cases.

%	 Each dataset includes 6 guided-movement sequences—comprising 3 translational movements along the X, Y, and Z axes and 3 rotational movements (roll, yaw, and pitch)—as well as 6 free-movement sequences, where subjects move their heads without constraints. All videos begin and end with the subject facing the camera at a distance of 55–60 cm. Translational movements extend up to 200 mm in any direction, while rotational movements reach a maximum of 30°. These constraints are consistently maintained across the entire dataset to ensure uniformity.
%	In this section, we present two examples: deterministic case of EKF and UKF for 3D facial tracking data, while the other is the stochastic case of EKF and UKF for 3D facial tracking data. The simulations in this section were conducted in Python. The key numerical results demonstrate that the EKF effectively reduces noise, resulting in a significant improvement in tracking accuracy in both examples. The database comprises 120 videos across 10 subjects (6 males and 4 females), with each subject having 12 videos. Each set contains 6 guided-movement sequences (3 translations in X, Y, Z, and 3 rotations in roll, yaw, pitch) and 6 free-movement sequences, where the user moves the head freely. All videos start and end with the head in a frontal position at a distance of 55–60 cm from the camera. Movements in the translations can range up to 200 mm in any direction, and rotations can reach up to 30°. For consistency, these constraints are maintained throughout the database.

\subsection{Example 1: Deterministic case of EKF and UKF for 3D facial tracking data }

In this example, we apply both the EKF and UKF models, as described in Section \ref{model-description}, in a deterministic setting to track a 3D facial model over time. To simplify the problem, we eliminate the stochastic components, such as process and measurement noise, from both the EKF and UKF models. The 3D facial tracking model is represented by a set of points in three-dimensional space, each defined by its $x, y$ and $z$ coordinates. The goal is to track these points across multiple time steps to estimate the evolving state of the model.

\subsubsection{State representation}

For the 3D facial tracking problem, the state vector \( \mathbf{x}_k \) at time step \( k \) represents the set of 3D coordinates of all \( N \) facial points:

\[
\mathbf{x}_k = [x_1(k), y_1(k), z_1(k), \dots, x_N(k), y_N(k), z_N(k)]^T \in \mathbb{R}^{3N},
\]
where \( N \) is the number of facial points (e.g., 54 facial points in this example), while each point has three coordinates \( (x_i(k), y_i(k), z_i(k)) \).

%The state at time \( t \) is represented as a vector containing the 3D coordinates of all facial points:
%
%\[
%x_t = \begin{bmatrix}
%	x_1, y_1, z_1, x_2, y_2, z_2, \dots, x_N, y_N, z_N
%\end{bmatrix}^T,
%\]
%where \( N \) is the number of 3D facial points. Each facial point has 3 coordinates (X, Y, Z), giving us a total state dimension of \( 3N \).

\subsubsection{Process model}

In this deterministic case, we assume the state evolves over time without any process noise. The state transition model is given by:

\[
\mathbf{x}_{t+1} = f(\mathbf{x}_t) = \mathbf{x}_t.
\]
%\[
%\mathbf{x}_{k+1} = \mathbf{x}_k + \mathbf{v}_k \Delta t,
%\]

This means that the state remains constant from one time step to the next, as no external forces or noise influence the state in this particular example.

\subsubsection{Measurement model}

The measurement model assumes that the observed state is identical to the true state (i.e., the measurements directly correspond to the coordinates of the facial points):

\[
\mathbf{z}_t = h(\mathbf{x}_t) = \mathbf{x}_t,
\]
where \( \mathbf{z}_t \) represents the measurement at time \( t \), and \( h(\mathbf{x}_t) \) is the identity function in this case, meaning the state is directly observed.

\subsection*{\bf EKF}
In the EKF model, we apply the algorithm provided in Section \ref{model-description}, where we remove the stochastic terms.
%The EKF algorithm is used to estimate the state of the system by predicting the next state and updating the estimate based on the incoming measurements. The process consists of two main steps: prediction and update.
%
%\subsubsection*{Prediction Step:}
%
%In the prediction step, the state is projected forward using the process model:
%
%\[
%\mathbf{\hat{x}}_{t+1|t} = f(\mathbf{x}_t)
%\]
%
%Since the process model is deterministic (i.e., \( f(\mathbf{x}_t) = \mathbf{x}_t \)), the predicted state is simply the previous state. The error covariance matrix \( P_{t+1|t} \) is updated as follows:
%
%\[
%P_{t|t-1} = F_{t-1} P_{t-1|t-1} F_{t-1}^T
%\]
%where \( F \) is the Jacobian of the process model. In this deterministic case, \( F \) is the identity matrix, so the covariance matrix remains unchanged unless modified externally.
%
%\subsubsection*{Update Step:}
%
%In the update step, the filter compares the predicted state \( \hat{x}_{t+1|t} \) to the new measurements \( z_t \). The measurement residual \( y_t \) is calculated:
%
%\[
%y_t = z_t - h(\hat{x}_{t+1|t}) = z_t - \hat{x}_{t+1|t}
%\]
%
%The Kalman gain \( K_t \) is computed as:
%
%\[
%K_t = P_{t+1|t} H^T (H P_{t+1|t} H^T + R)^{-1}
%\]
%
%Where \( H \) is the measurement matrix (which is the identity matrix in this case), and \( R \) is the measurement noise covariance (assumed to be zero in this deterministic example).
%
%The state estimate is updated as:
%
%\[
%\hat{x}_{t+1} = \hat{x}_{t+1|t} + K_t y_t
%\]
%
%The error covariance matrix is updated as:
%
%\[
%P_{t+1} = (I - K_t H) P_{t+1|t}
%\]

\subsection*{\bf UKF}

The UKF is an alternative to the EKF, which uses a set of sigma points to better approximate the state distribution, particularly in cases where the state distribution is highly nonlinear. However, in this deterministic example, the UKF performs similarly to the EKF since there is no process noise or nonlinearity. Here, we apply the UKF algorithm provided in Section \ref{model-description}, where we remove the stochastic terms.

%\subsubsection*{Sigma Point Generation:}
%
%
%Sigma points are generated from the current state estimate and the error covariance matrix. In the deterministic case, the sigma points are computed as:
%
%\[
%\chi_0 = \hat{x}
%\]
%
%\[
%\chi_i = \hat{x} + (\sqrt{(n + \lambda)P})_i \quad \text{for} \quad i = 1, \dots, n
%\]
%
%Where \( n \) is the state dimension, \( P^{1/2} \) is the square root of the covariance matrix, and \( \lambda \) is a scaling parameter.
%
%\subsubsection*{Prediction Step:}
%
%The sigma points are propagated through the process model. Since the process model is deterministic (i.e., \( f(x_t) = x_t \)), the predicted sigma points are the same as the original sigma points:
%
%\[
%\hat{x}_{t+1|t} = \sum_{i=0}^{2n} W_m^{(i)} \chi_i
%\]
%
%Where \( W_m^{(i)} \) are the weights for the mean, which are computed based on the sigma points.
%
%\subsubsection*{Update Step:}
%
%The predicted measurement is calculated as the weighted mean of the measurements corresponding to the propagated sigma points:
%
%\[
%\hat{z}_{t+1|t} = \sum_{i=0}^{2n} W_m^{(i)} h(\chi_i)
%\]
%
%The Kalman gain \( K_t \) is computed using the predicted measurement and the predicted state covariance:
%
%\[
%K_t = P_{xz} S_t^{-1},
%\]
%where \( P_{xz} \) is the cross-covariance between the state and the measurement, and \( S_t \) is the innovation covariance.
%
%The state is updated as:
%
%\[
%\hat{x}_{t+1} = \hat{x}_{t+1|t} + K_t (z_t - \hat{z}_{t+1|t})
%\]
%
%The error covariance matrix is updated as:
%
%\[
%P_{t+1} = P_{t+1|t} - K_t S_t K_t^T
%\]

\begin{figure}[h!]
	\centering
	\begin{tabular}{ll}
		\includegraphics[width=0.45\textwidth]{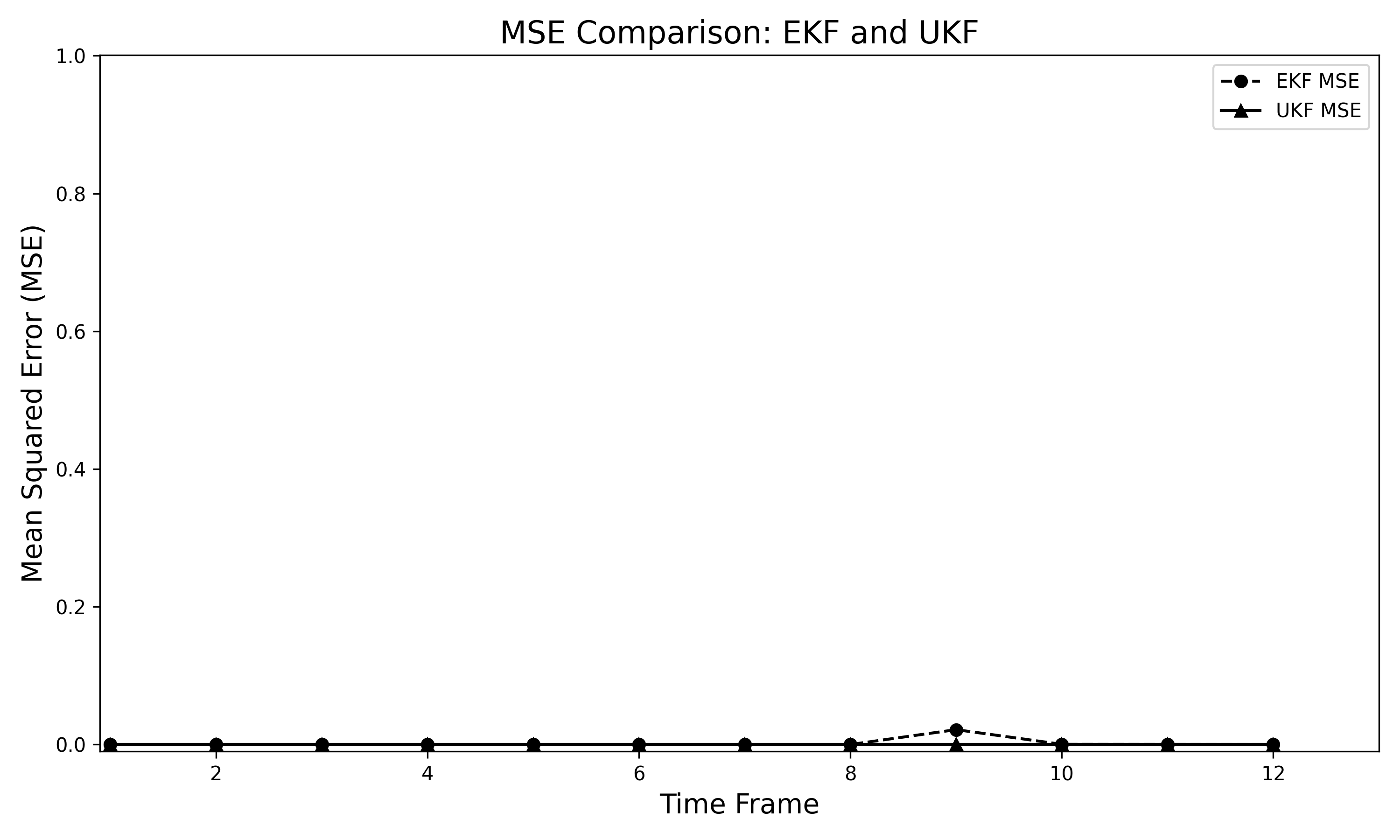} &\includegraphics[width=0.45\textwidth]{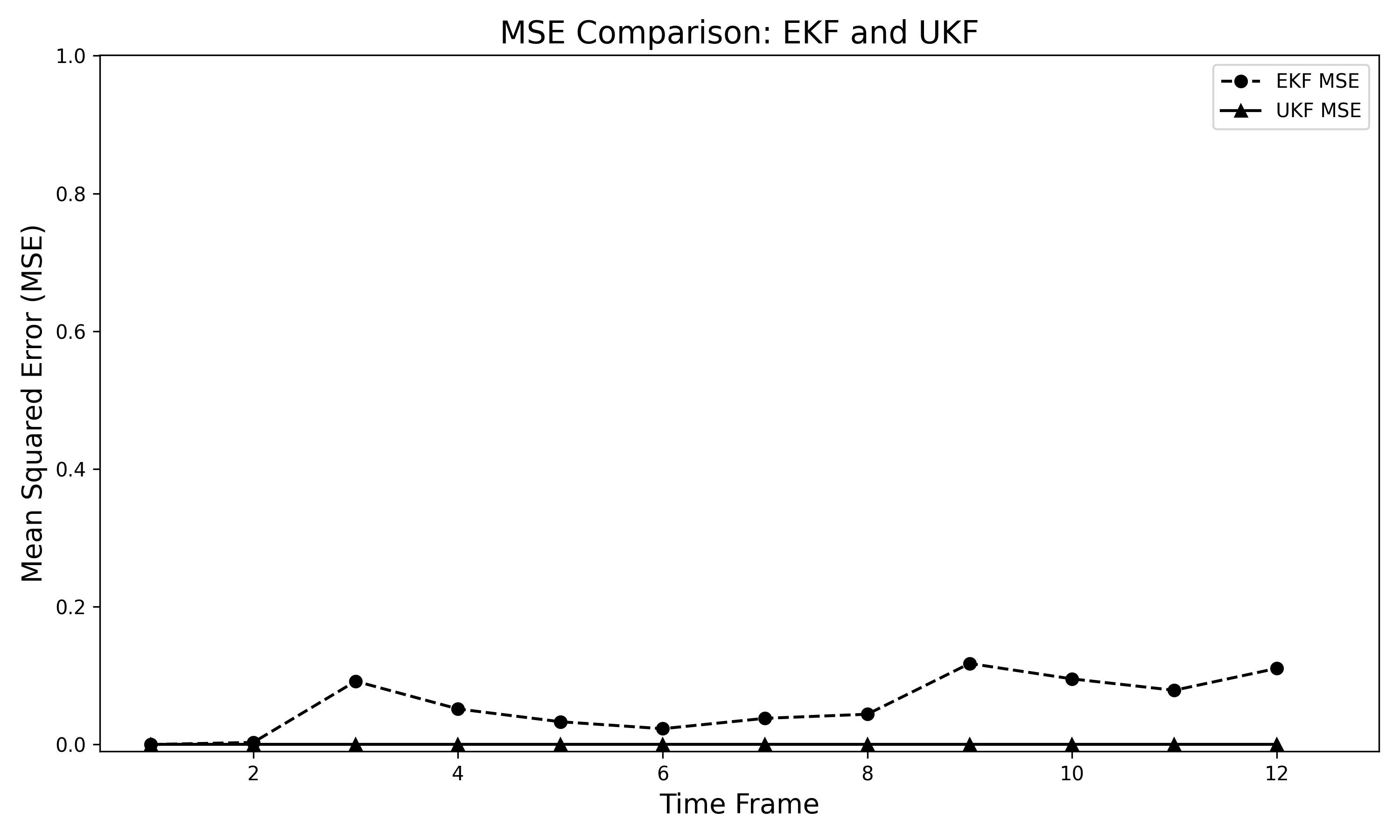} \\
		\includegraphics[width=0.45\textwidth]{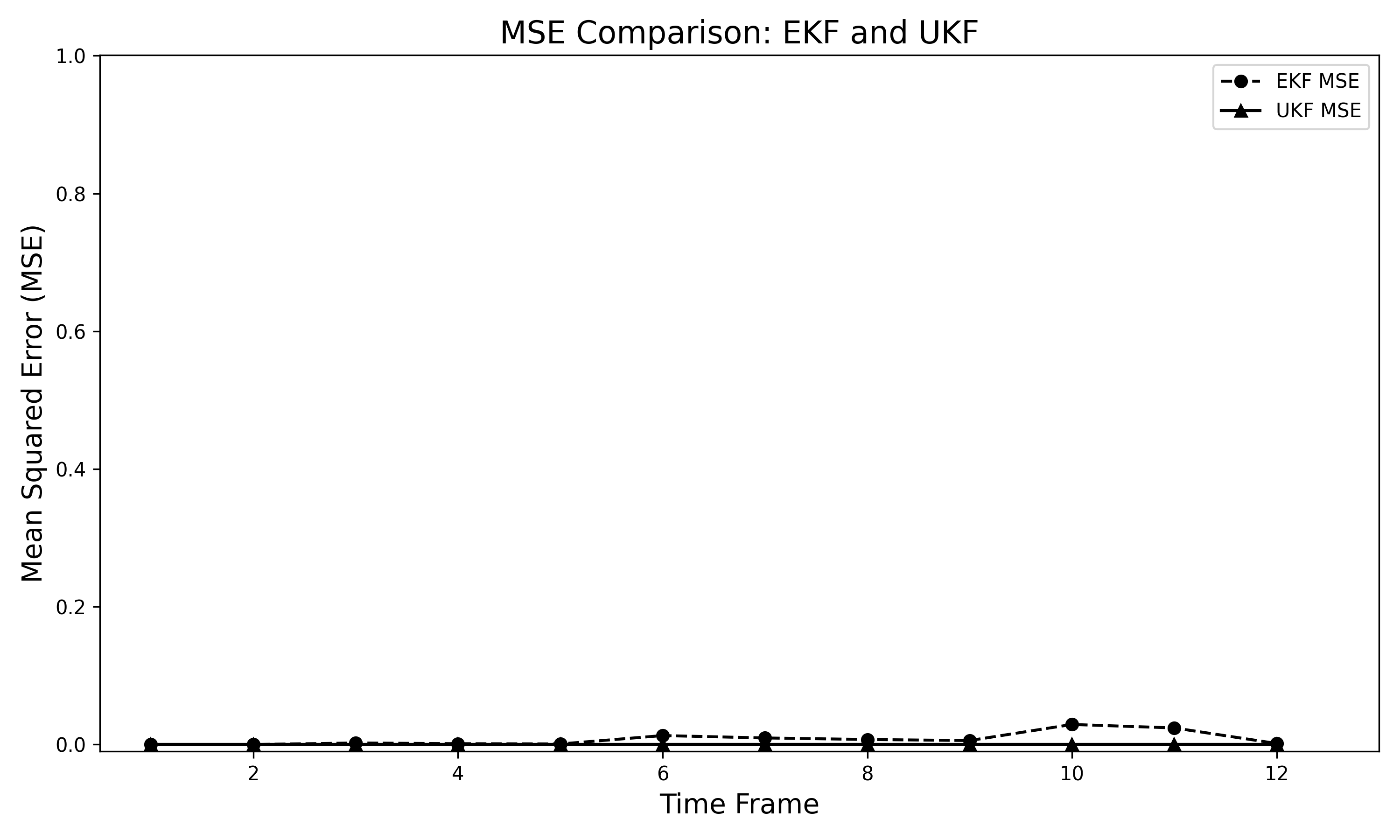} &\includegraphics[width=0.45\textwidth]{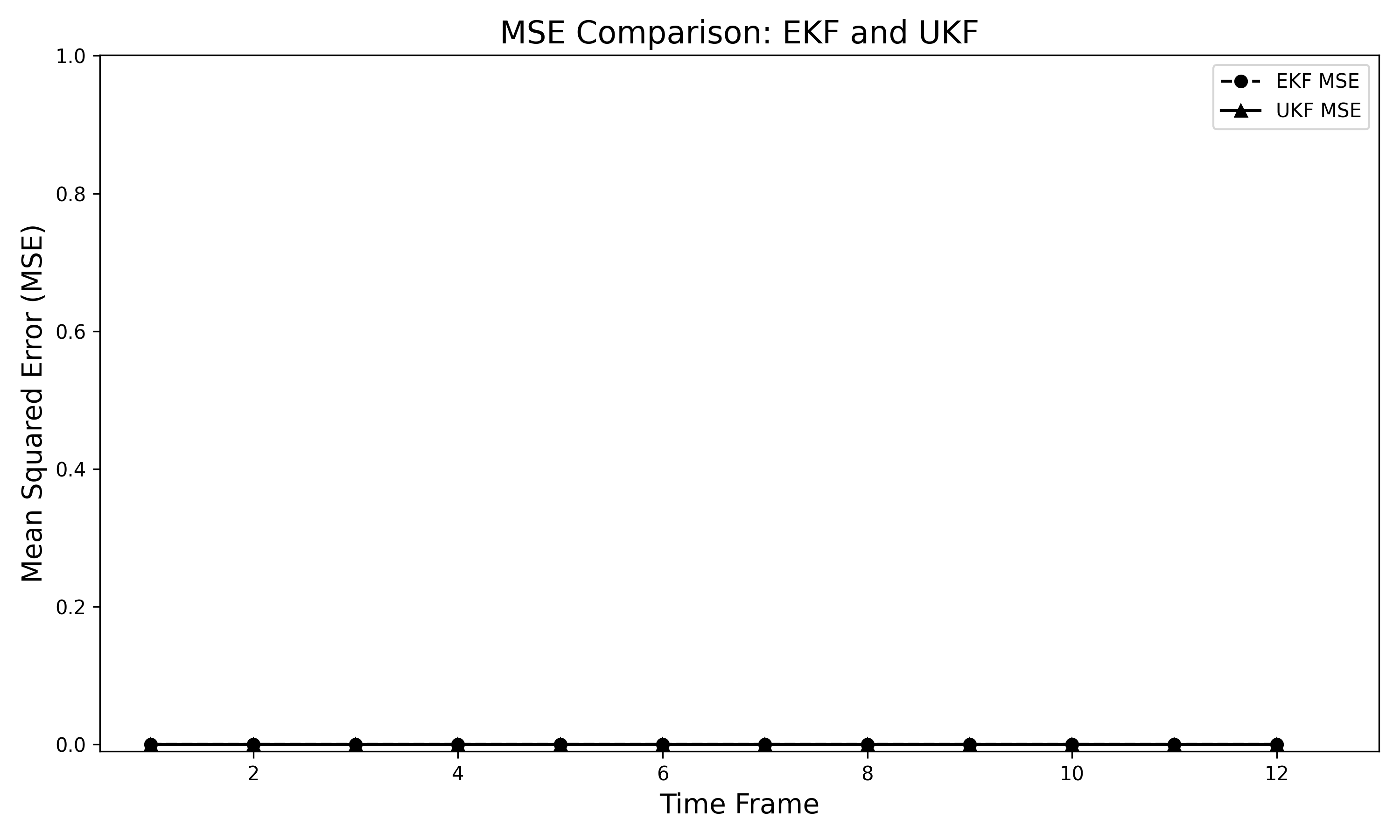}
	\end{tabular}
	\caption{[Color online] Comparison of the MSE between the EKF and UKF for deterministic 3D facial motion tracking over 12 time-frames for users 1,2,3,4 (panels from the left to the right and from the top to the bottom). The experiment tracks 54 facial points (each with X, Y, and Z coordinates), using a deterministic process model with no process noise and a direct measurement model. The UKF, leveraging sigma points for nonlinear state estimation, demonstrates lower MSE compared to the EKF, which relies on linearization. The dataset used in this simulation is provided in  \cite{Ariz2016novel}, licensed under a Creative Commons Attribution-NonCommercial-ShareAlike 4.0 International License. }
%		The plot highlights the improved accuracy of UKF in reducing estimation errors, making it a more effective choice for precise motion tracking in applications such as facial recognition and animation.}
	\label{fig:3}
\end{figure}

\begin{figure}[h!]
	\centering
	\begin{tabular}{ll}
		\includegraphics[width=0.45\textwidth]{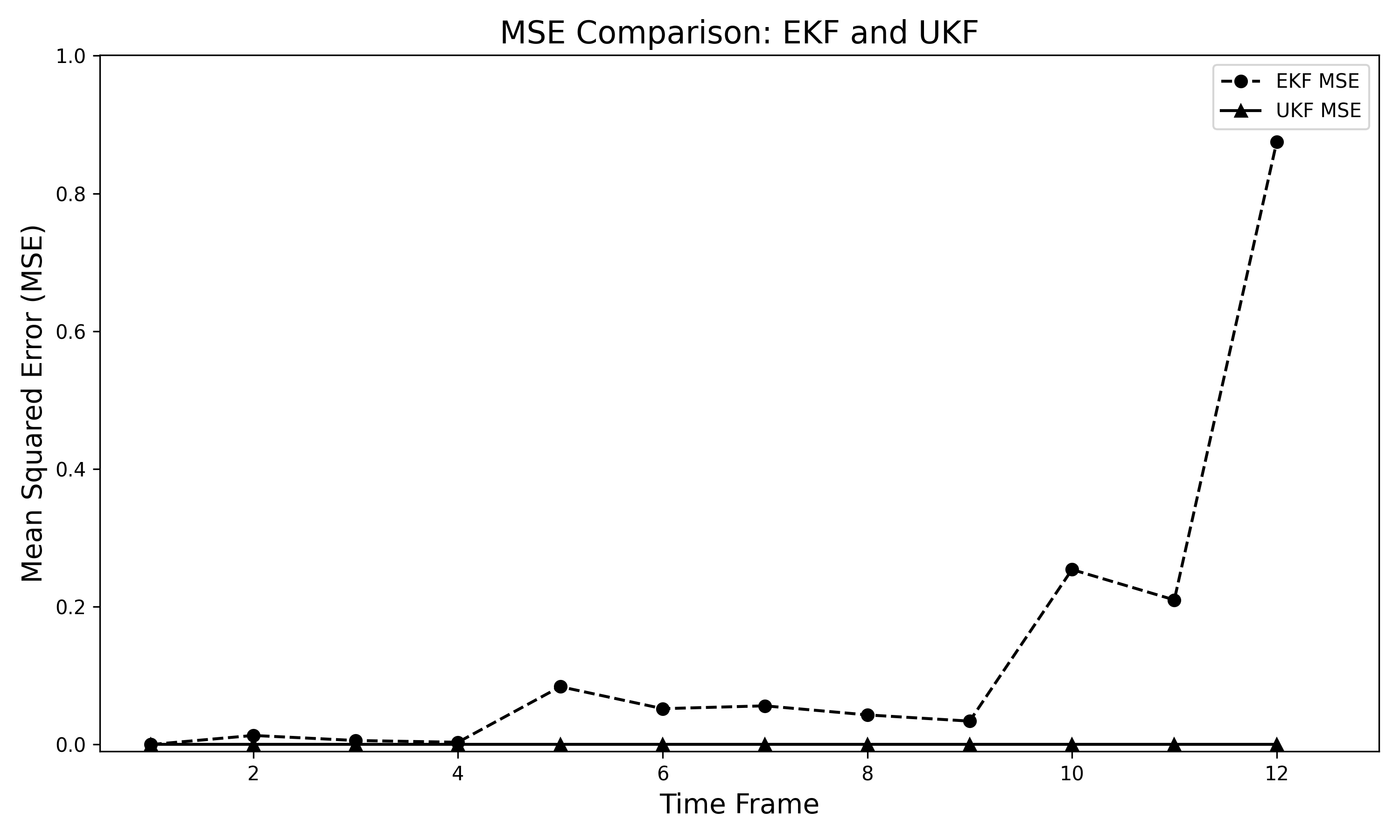} &\includegraphics[width=0.45\textwidth]{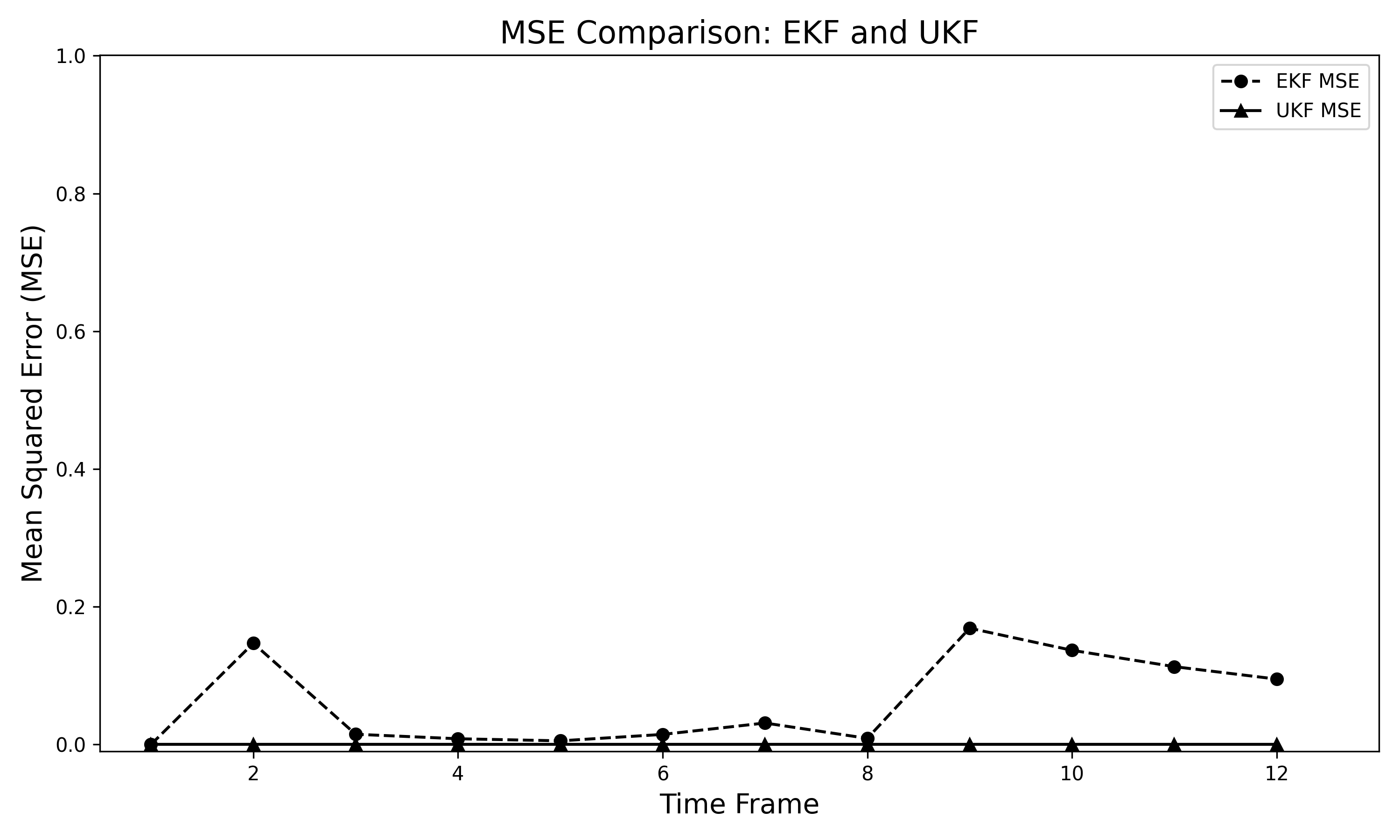} \\
		\includegraphics[width=0.45\textwidth]{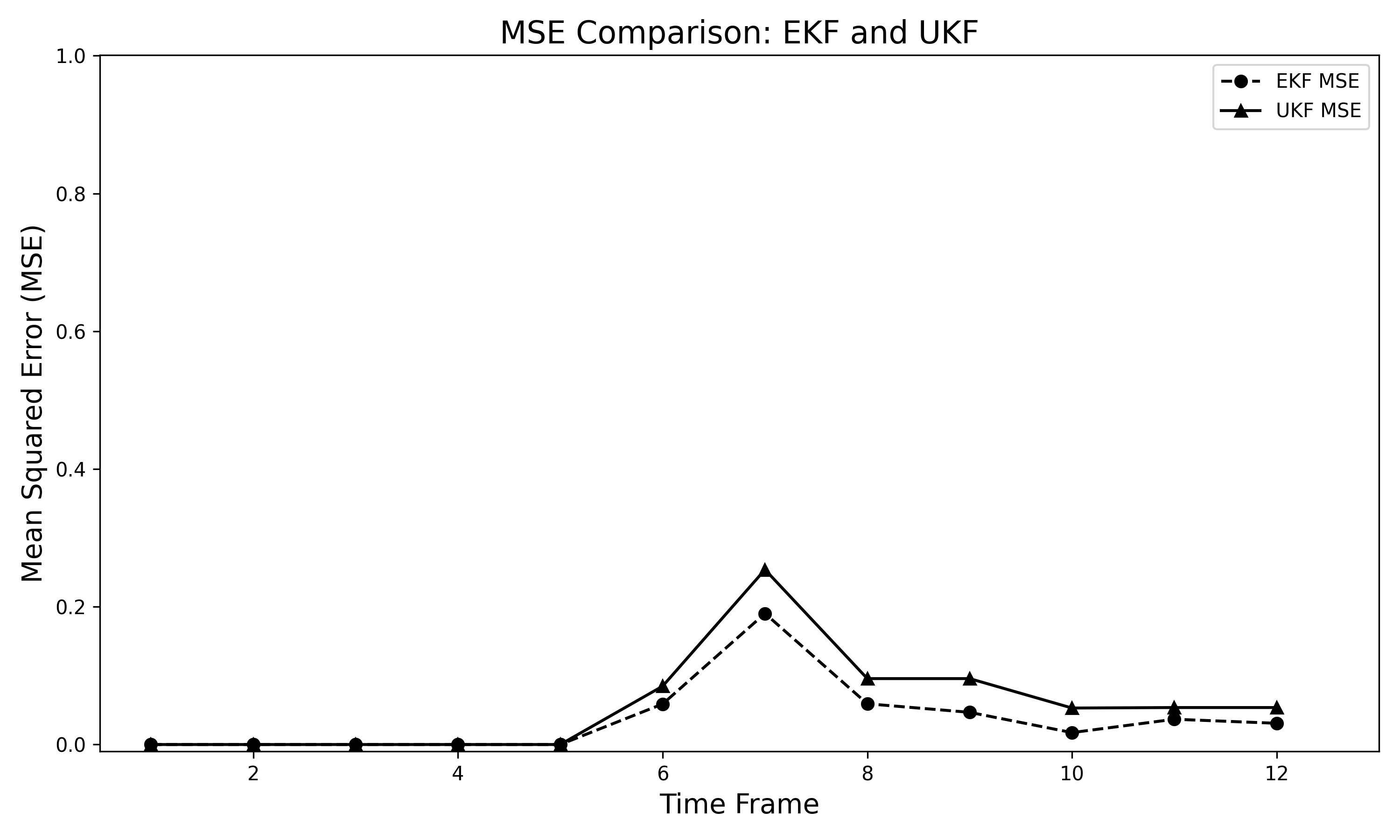} &\includegraphics[width=0.45\textwidth]{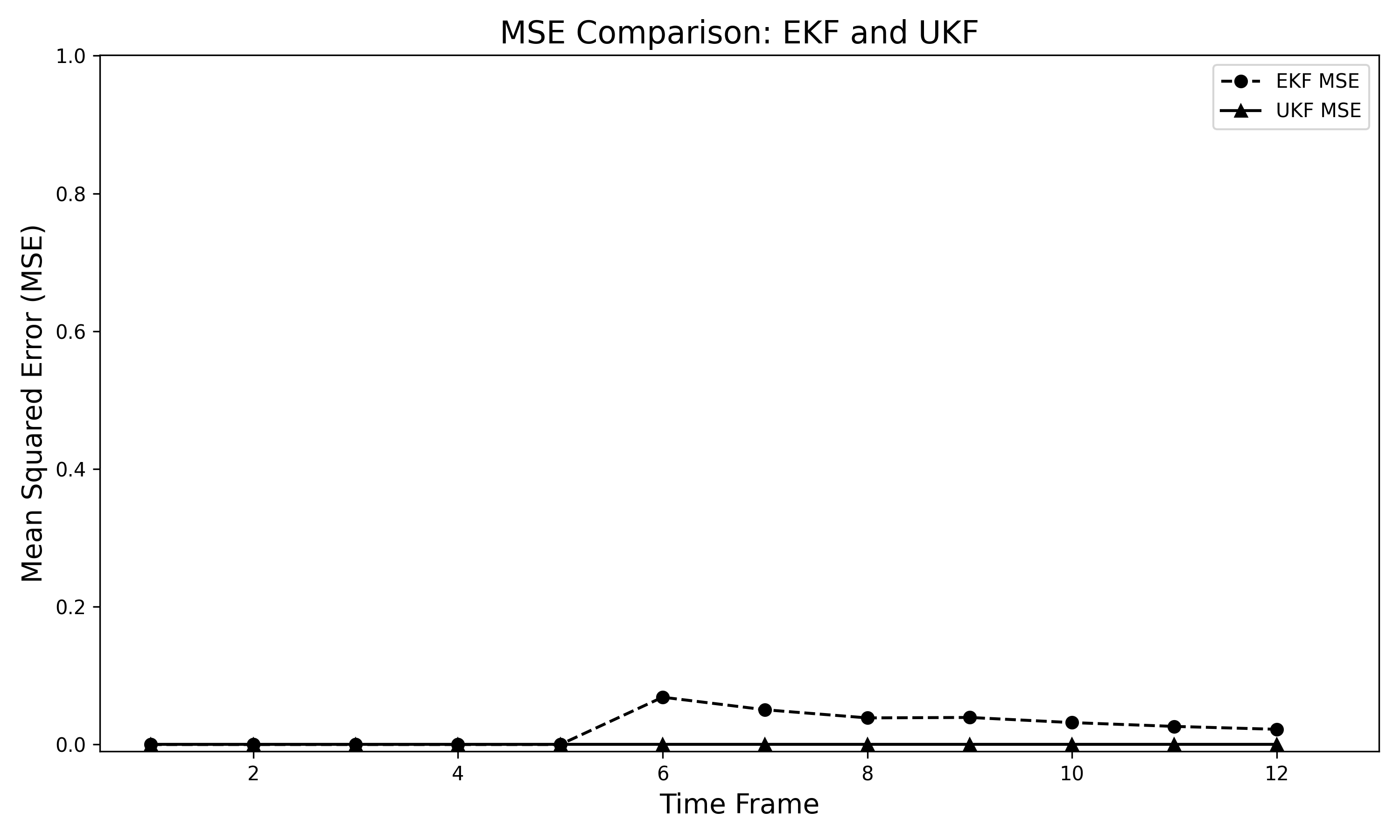}
	\end{tabular}
	\caption{[Color online] Comparison of the MSE between the EKF and UKF for deterministic 3D facial motion tracking over 12 time-frames for users 5,6,7,8 (panels from the left to the right and from the top to the bottom). The experiment tracks 54 facial points (each with X, Y, and Z coordinates), using a deterministic process model with no process noise and a direct measurement model. The UKF, leveraging sigma points for nonlinear state estimation, demonstrates lower MSE compared to the EKF, which relies on linearization. The dataset used in this simulation is provided in  \cite{Ariz2016novel},  licensed under a Creative Commons Attribution-NonCommercial-ShareAlike 4.0 International License. }
	\label{fig:4}
\end{figure}

\begin{figure}[h!]
	\centering
	\begin{tabular}{ll}
		\includegraphics[width=0.45\textwidth]{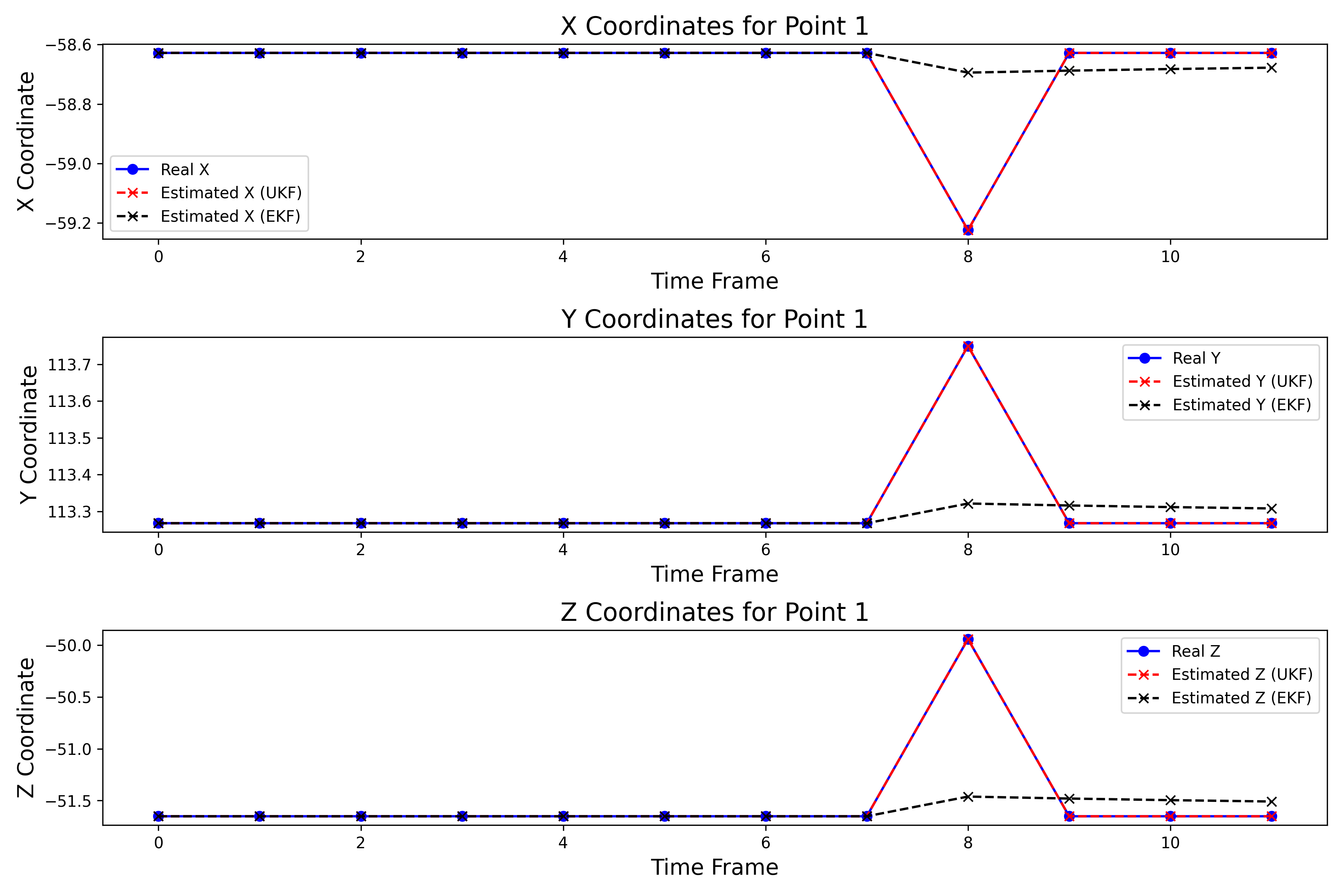} &\includegraphics[width=0.45\textwidth]{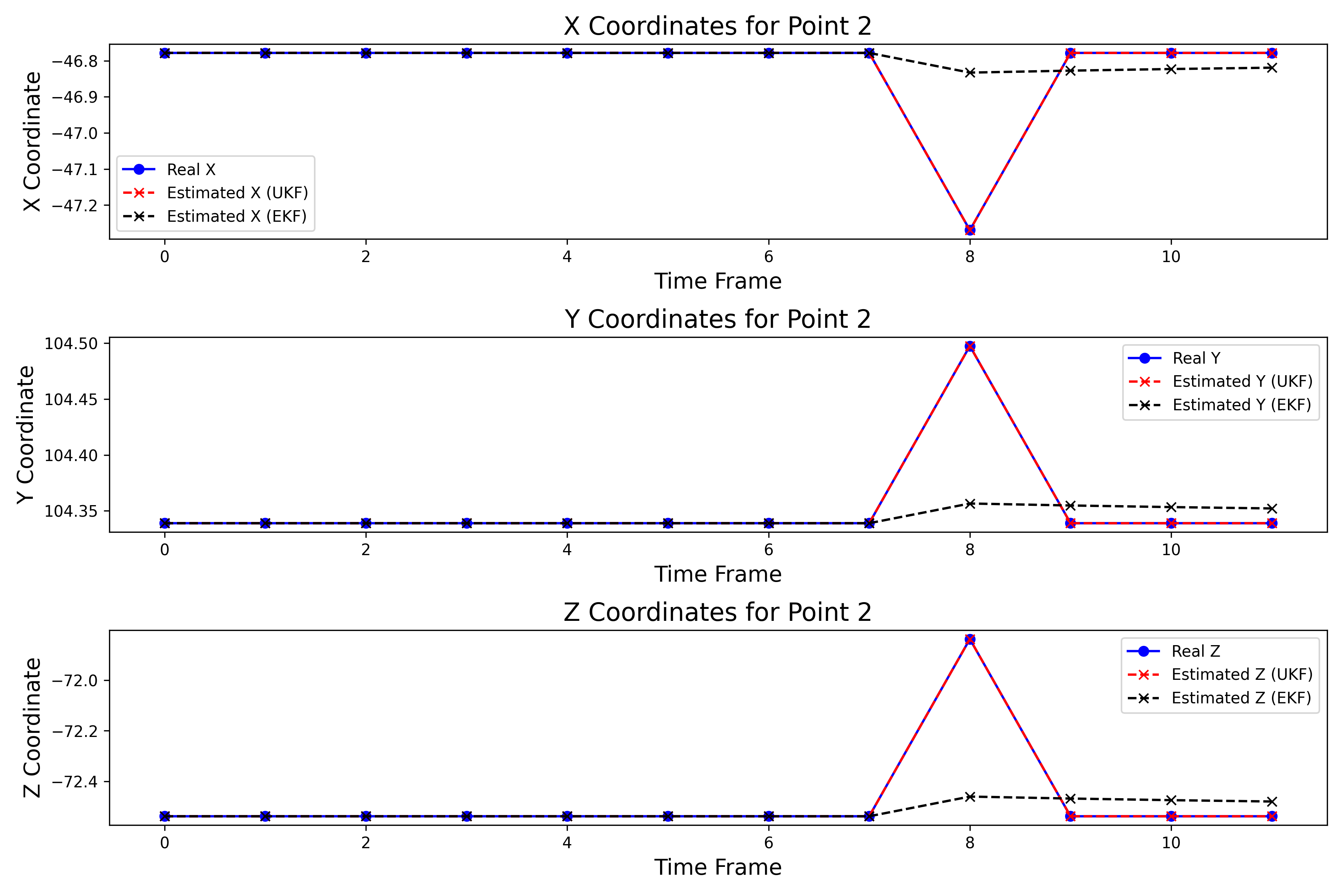} \\
		\includegraphics[width=0.45\textwidth]{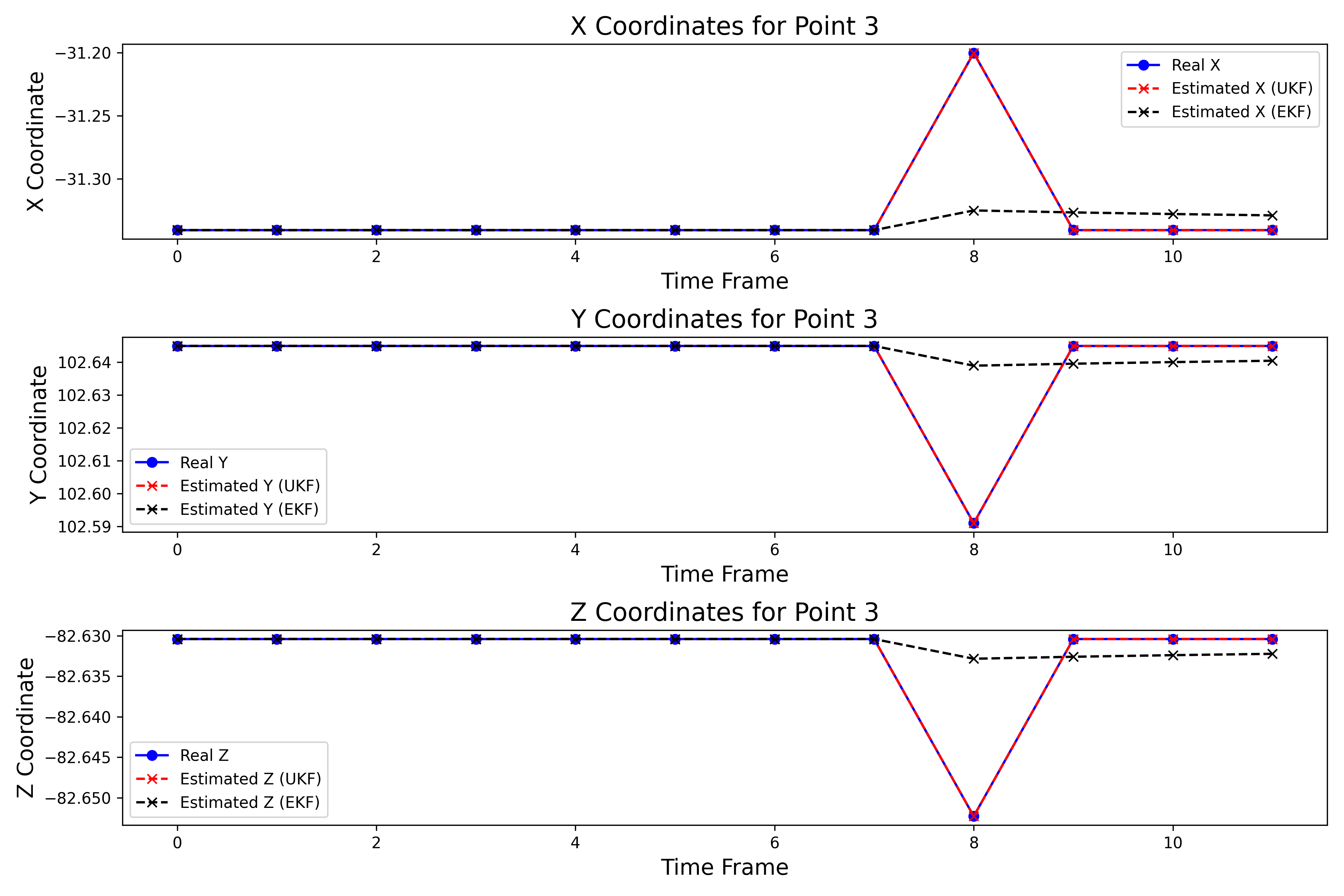} &\includegraphics[width=0.45\textwidth]{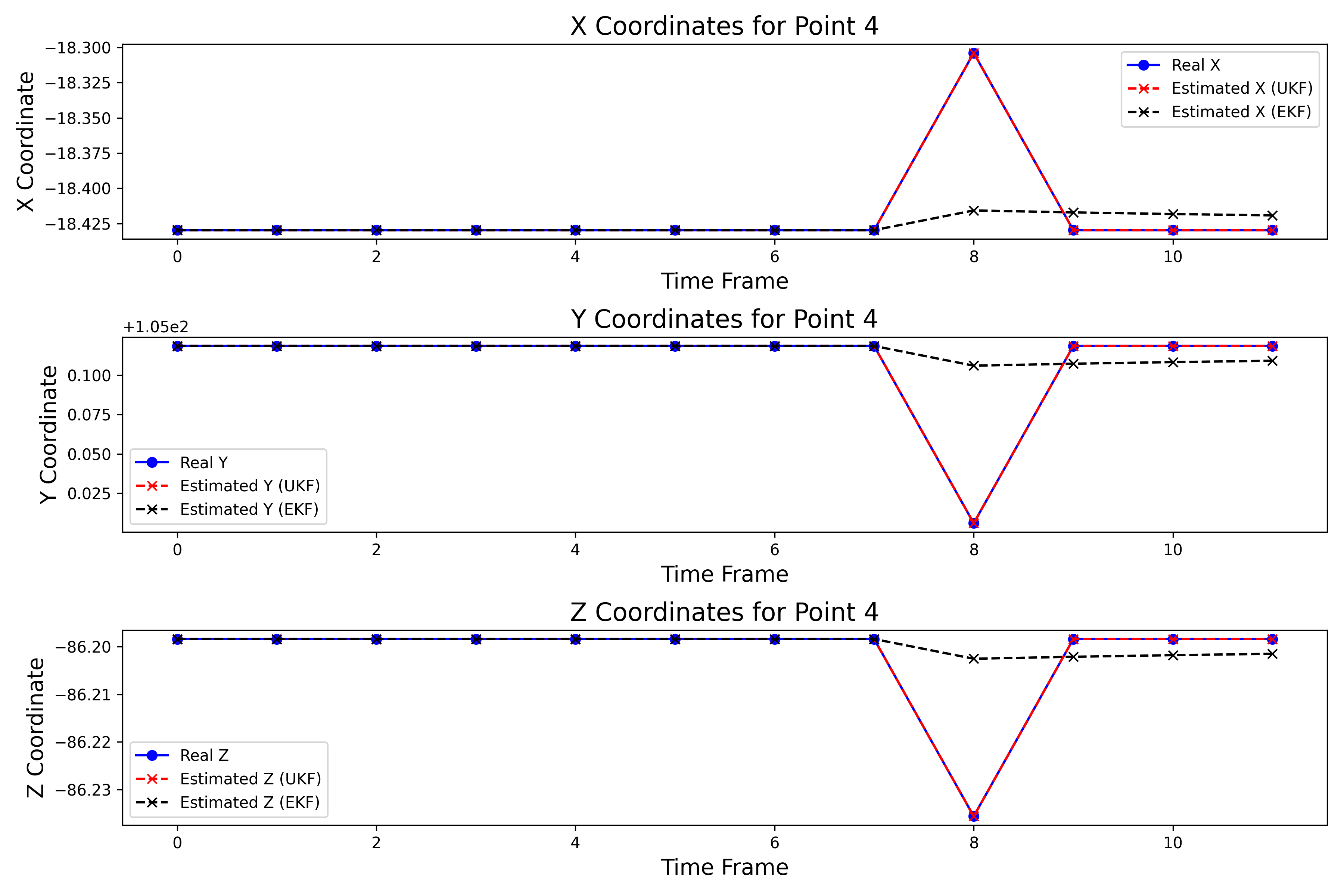}
	\end{tabular}
	\caption{[Color online] Comparison of real and estimated 3D facial landmark coordinates $x,y,z$ over the time frames for 4 selected points for user 1. The real data is plotted alongside the estimated values obtained using the deterministic UKF and EKF. Each subplot corresponds to a specific coordinate ($x, y$ or $z$), showing temporal variations across frames. The dataset used in this simulation is provided in  \cite{Ariz2016novel}, licensed under a Creative Commons Attribution-NonCommercial-ShareAlike 4.0 International License. }
	\label{fig:2}
\end{figure}

In a noise-free environment, in Figs. \ref{fig:2}-\ref{fig:4}, our presented results show that UKF consistently produces a smaller MSE than EKF. In particular, we see that the MSE of UKF and EKF are almost very close to 0. The MSE of UKF are smaller than the MSE of EKF in all cases. However, the MSE of UKF fluctuate to a value of approximately 0.9 in the top left panel in Fig. \ref{fig:4}.  The increase in UKF MSE may be due to the poor sigma point approximation in highly nonlinear regions, numerical instabilities, or an ill-conditioned covariance matrix. Additionally, sensitivity to initial conditions and boundary effects of certain facial landmarks may have contributed to this fluctuation.
\begin{figure}[h!]
	\centering
	\begin{tabular}{ll}
		\includegraphics[width=0.45\textwidth]{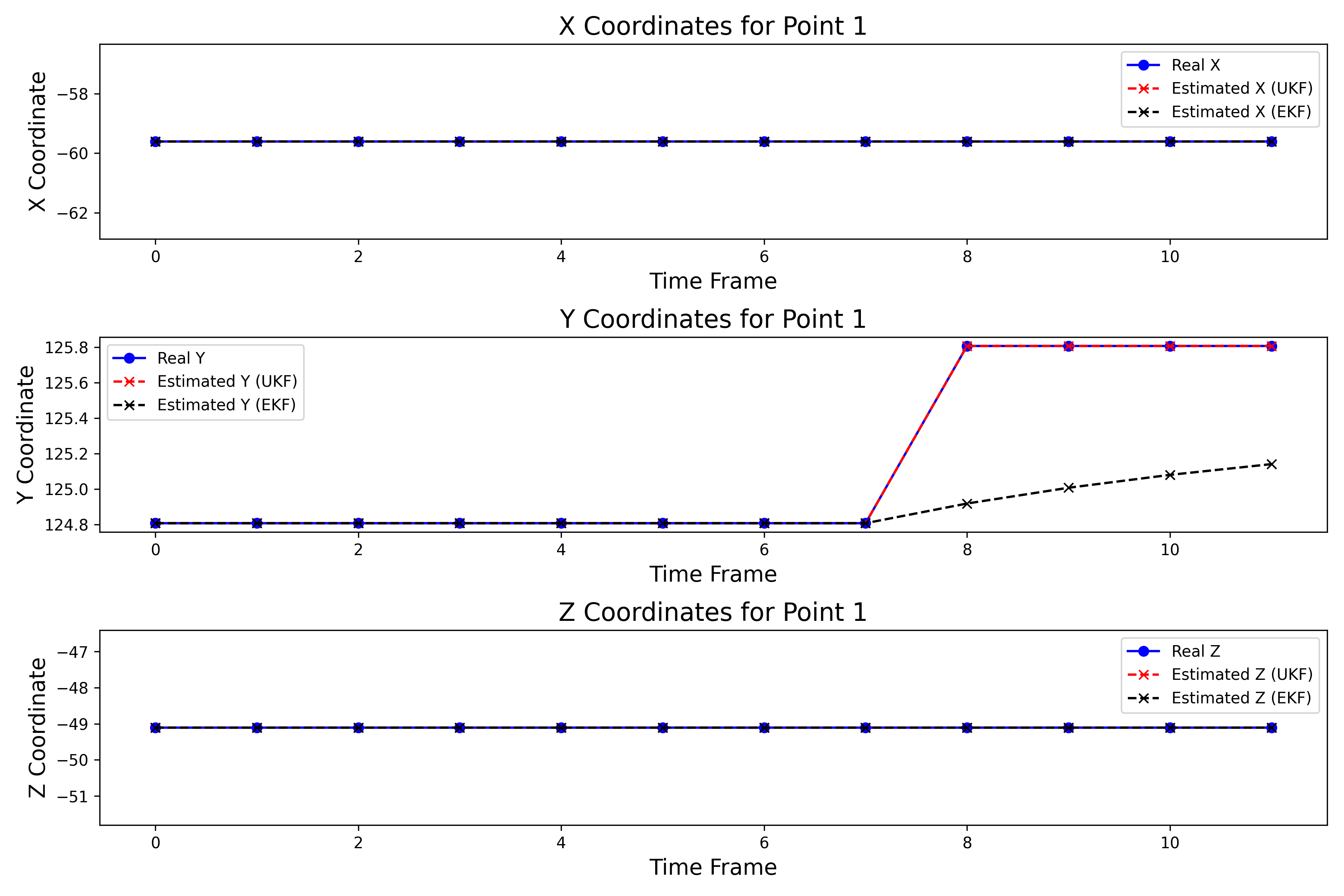} &\includegraphics[width=0.45\textwidth]{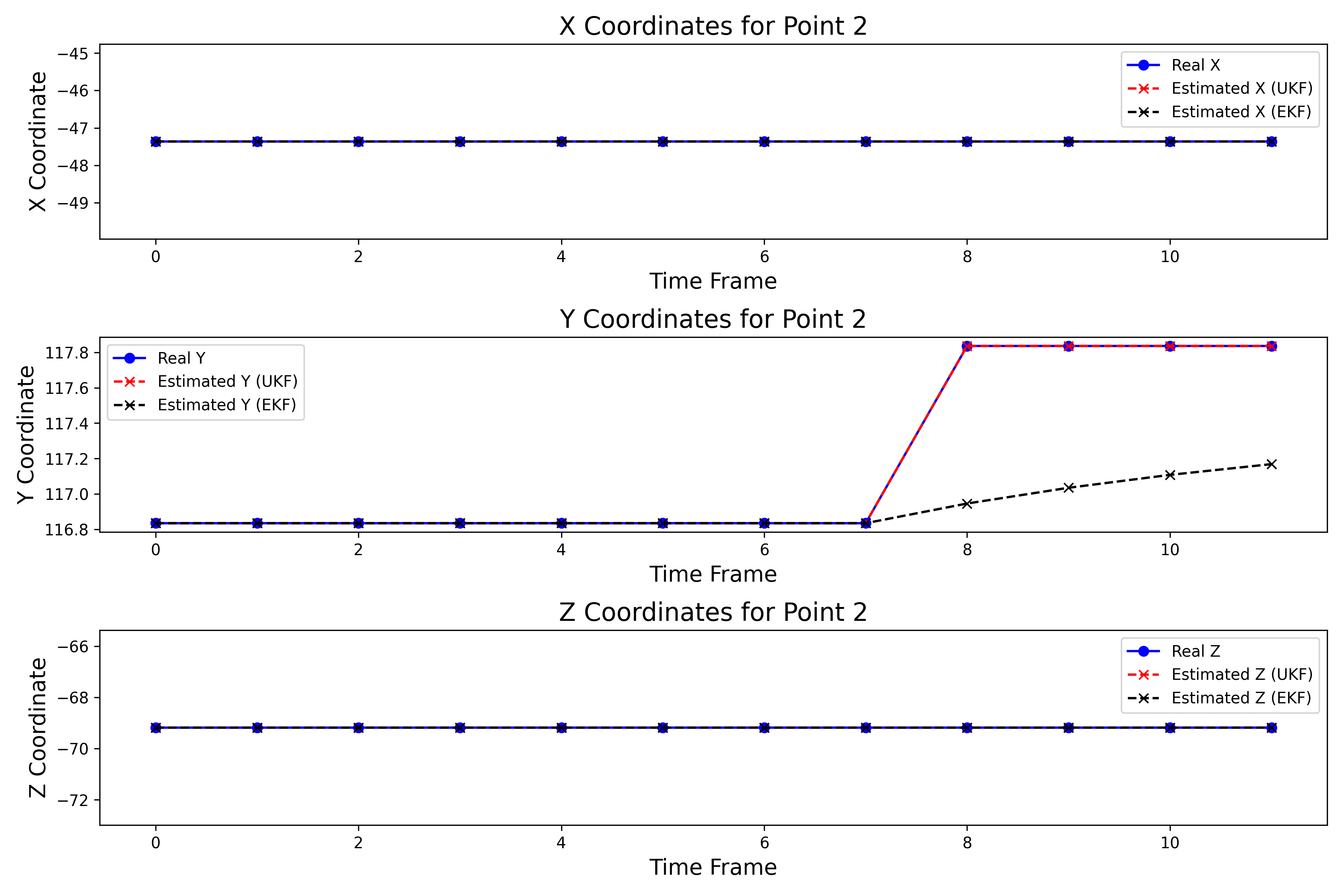} \\
		\includegraphics[width=0.45\textwidth]{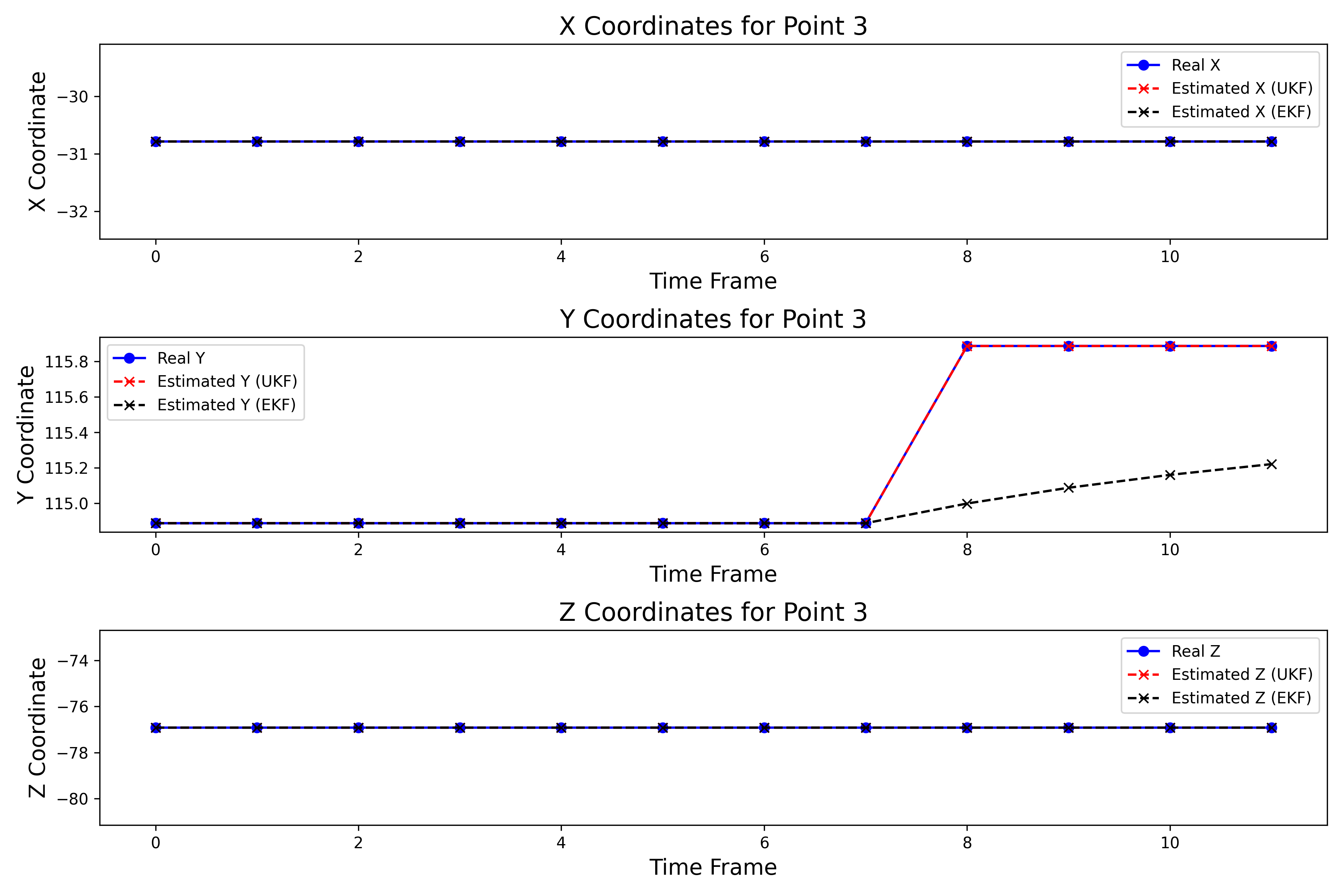} &\includegraphics[width=0.45\textwidth]{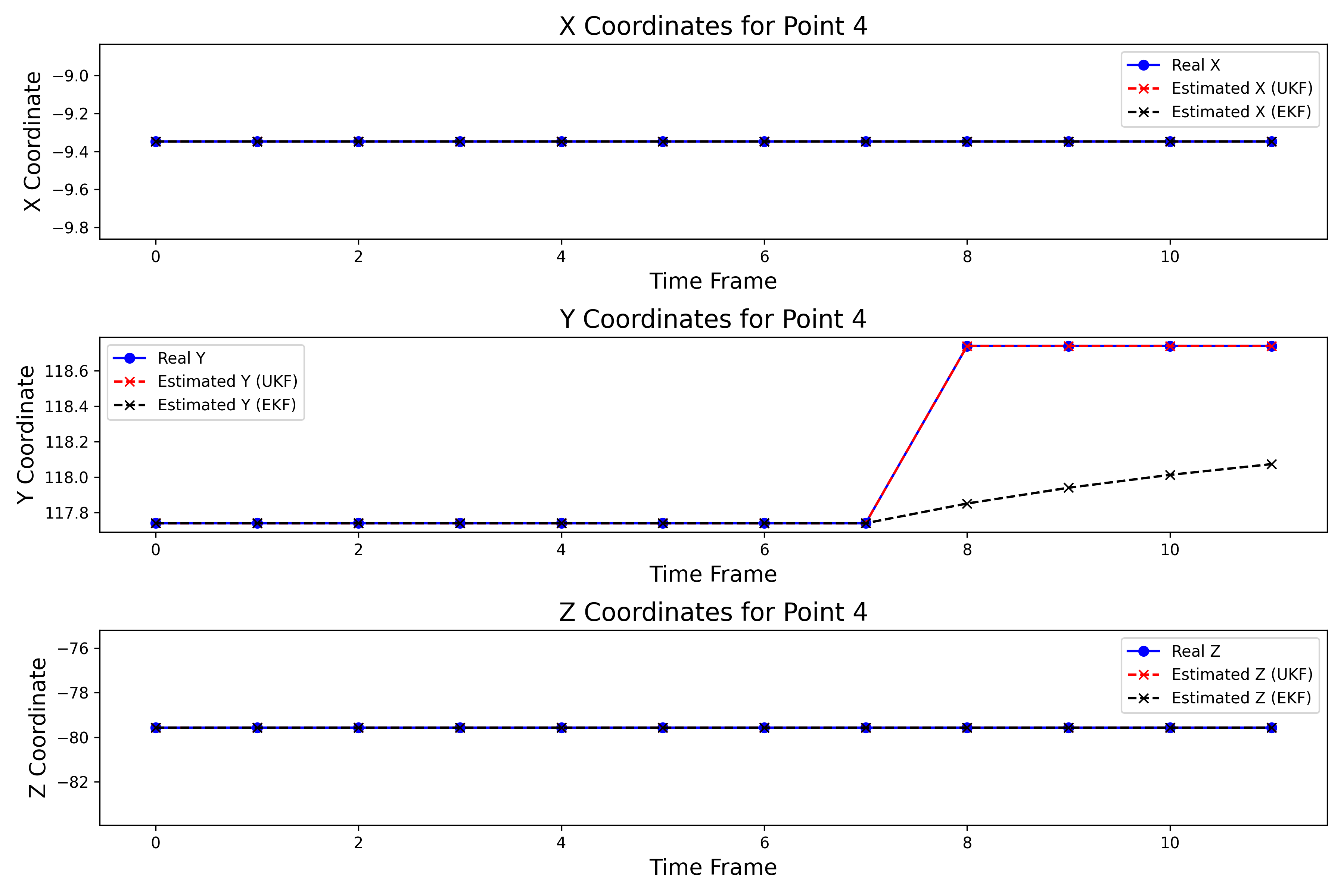}
	\end{tabular}
	\caption{[Color online] Comparison of real and estimated 3D facial landmark coordinates $x,y,z$ over the time frames for 4 selected points for user 2. The real data is plotted alongside the estimated values obtained using the deterministic UKF and EKF. Each subplot corresponds to a specific coordinate ($x, y$ or $z$), showing temporal variations across frames. The dataset used in this simulation is provided in  \cite{Ariz2016novel}, licensed under a Creative Commons Attribution-NonCommercial-ShareAlike 4.0 International License. }
	\label{fig:2-2}
\end{figure}

This outcome highlights UKF’s advantage in accurately capturing nonlinear facial dynamics. Why does UKF perform better in this case?
Unlike EKF, which uses first-order Taylor series approximation, UKF propagates sigma points through the nonlinear system, capturing higher-order statistics. This approach significantly reduces approximation errors. UKF provides a more precise estimate of facial motion, leading to improved tracking performance. Since noise is absent or negligible, UKF fully exploits its ability to track smooth, nonlinear facial movements with higher precision. Moreover, in Figs \ref{fig:2}-\ref{fig:2-2}, we observe that the UKF estimates fit the real data better than the EKF approximation  over the time frames for 4 selected points for users 1 and 2. This result aligns with the observation obtained in Figs. \ref{fig:3}-\ref{fig:4}. 

In practical applications such as 3D face recognition or facial motion capture, UKF is preferable in controlled environments where sensor noise is minimal. High-quality depth cameras, for instance, can provide clean data that maximizes UKF’s advantage over EKF.
%	#### **Conclusion**
%	By applying the EKF and UKF to track a 3D face model in this deterministic setting, we can compare the performance of both algorithms in terms of their accuracy in predicting the facial points over time. The error metrics, such as MAE and MSE, provide insights into the efficacy of each filter and help identify the best filtering technique for this type of tracking problem.
%	
%	---
%	
%	This explanation should provide a clear and concise mathematical description of the deterministic EKF and UKF algorithms, along with how they are applied to the 3D face model tracking problem. You can use this as a section in your article to demonstrate the mathematical foundation of the filters.

\subsection{Example 2: Stochastic case of EKF and UKF for 3D facial tracking data} 
In this example, we apply both the EKF and UKF models, as described in Section \ref{model-description}, to track a 3D facial model over time. Similar to Example 1, the 3D facial tracking model consists of a set of points in three-dimensional space, each represented by its $x, y$ and $z$ coordinates. The goal is to track these points across multiple time steps to estimate the evolving state of the model. However, unlike Example 1, where a deterministic setting was assumed, here we consider stochastic models, incorporating process and measurement noise to account for the inherent uncertainty in real-world measurements.
\subsection*{ State Definition for 3D Facial Tracking}

For the 3D facial tracking problem, the state vector \( \mathbf{x}_k \) at time step \( k \) represents the set of 3D coordinates of all \( N \) facial points:

\[
\mathbf{x}_k = [x_1(k), y_1(k), z_1(k), \dots, x_N(k), y_N(k), z_N(k)]^T \in \mathbb{R}^{3N},
\]

where \( N \) is the number of facial points (e.g., 54 facial points in this example), while each point has three coordinates \( (x_i(k), y_i(k), z_i(k)) \).

%\subsection*{ Process Model (State Evolution)}

The state evolution follows a simple motion model with random velocity and Gaussian process noise:

\[
\mathbf{x}_{k+1} = \mathbf{x}_k + \mathbf{v}_k \Delta t + \mathbf{w}_k,
\]
where \( \mathbf{v}_k \) is the velocity of the facial points at time step \( k \), modeled as random noise

\[
\mathbf{v}_k \sim \mathcal{N}(0, \sigma_{\text{velocity}}^2 \mathbf{I}_{3N}).
\]

 Moreover, \( \Delta t \) is the time step between frames (e.g., \( 0.01 \) seconds) and  \( \mathbf{w}_k \) is the process noise vector

\[
\mathbf{w}_k \sim \mathcal{N}(0, \mathbf{Q}).
\]

%\subsection*{Measurement Model}

The measurement model assumes that the state is directly observed (identity model) with added Gaussian measurement noise:

\[
\mathbf{z}_k = \mathbf{H} \mathbf{x}_k + \mathbf{v}_k
\]  
where $\mathbf{H}= \mathbf{I}_{3N}$ is the identity matrix, \( \mathbf{z}_k \in \mathbb{R}^{3N} \) is the noisy measurement at time step \( k \), while \( \mathbf{v}_k \in \mathbb{R}^{3N} \) is the Gaussian measurement noise:

\[
\mathbf{v}_k \sim \mathcal{N}(0, \mathbf{R}).
\]

The measurement noise covariance \( \mathbf{R} \) is assumed to be diagonal, i.e.,

\[
\mathbf{R} = \sigma_{\text{measurement}}^2 \mathbf{I}_{3N}.
\]

\begin{figure}[h!]
	\centering
	\begin{tabular}{ll}
		\includegraphics[width=0.45\textwidth]{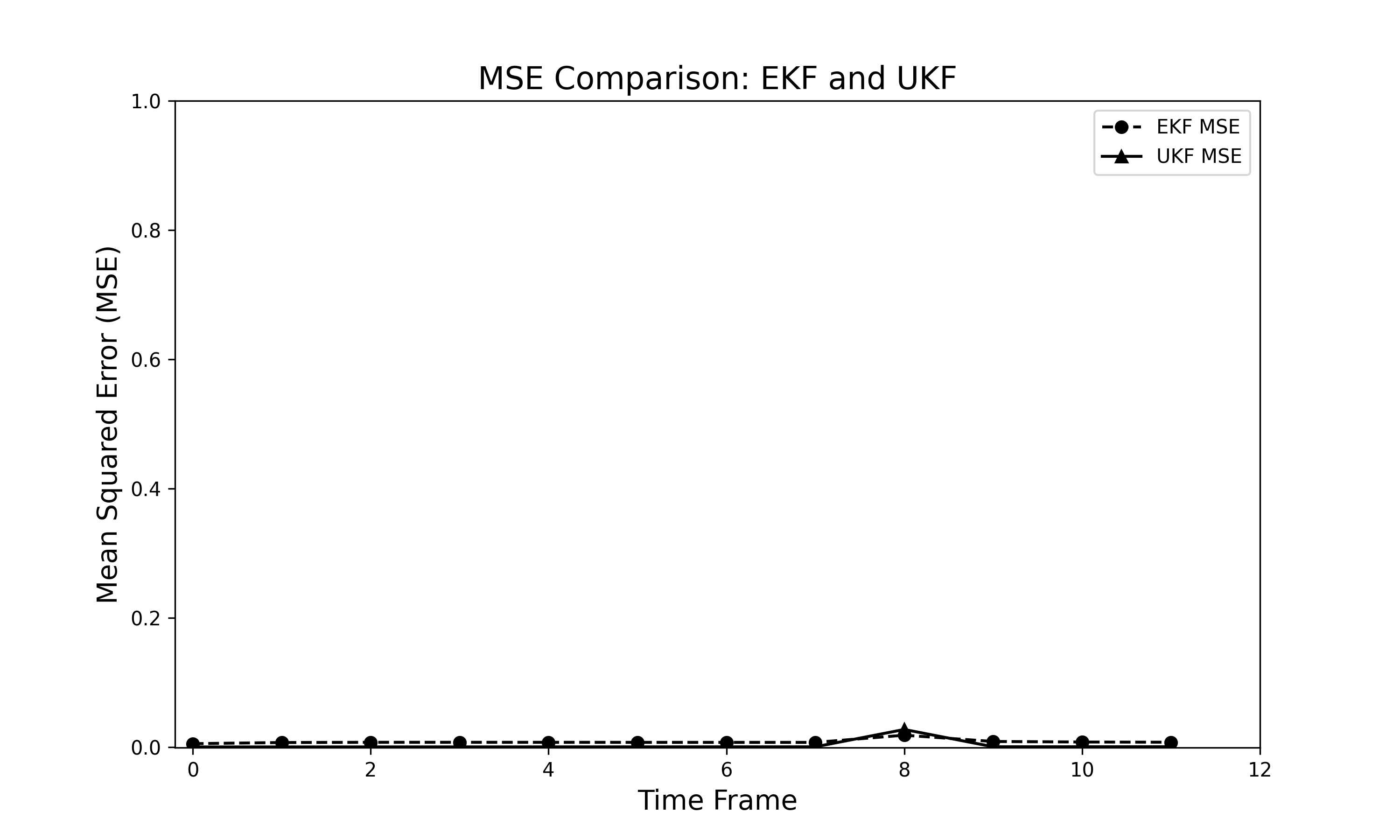} &\includegraphics[width=0.45\textwidth]{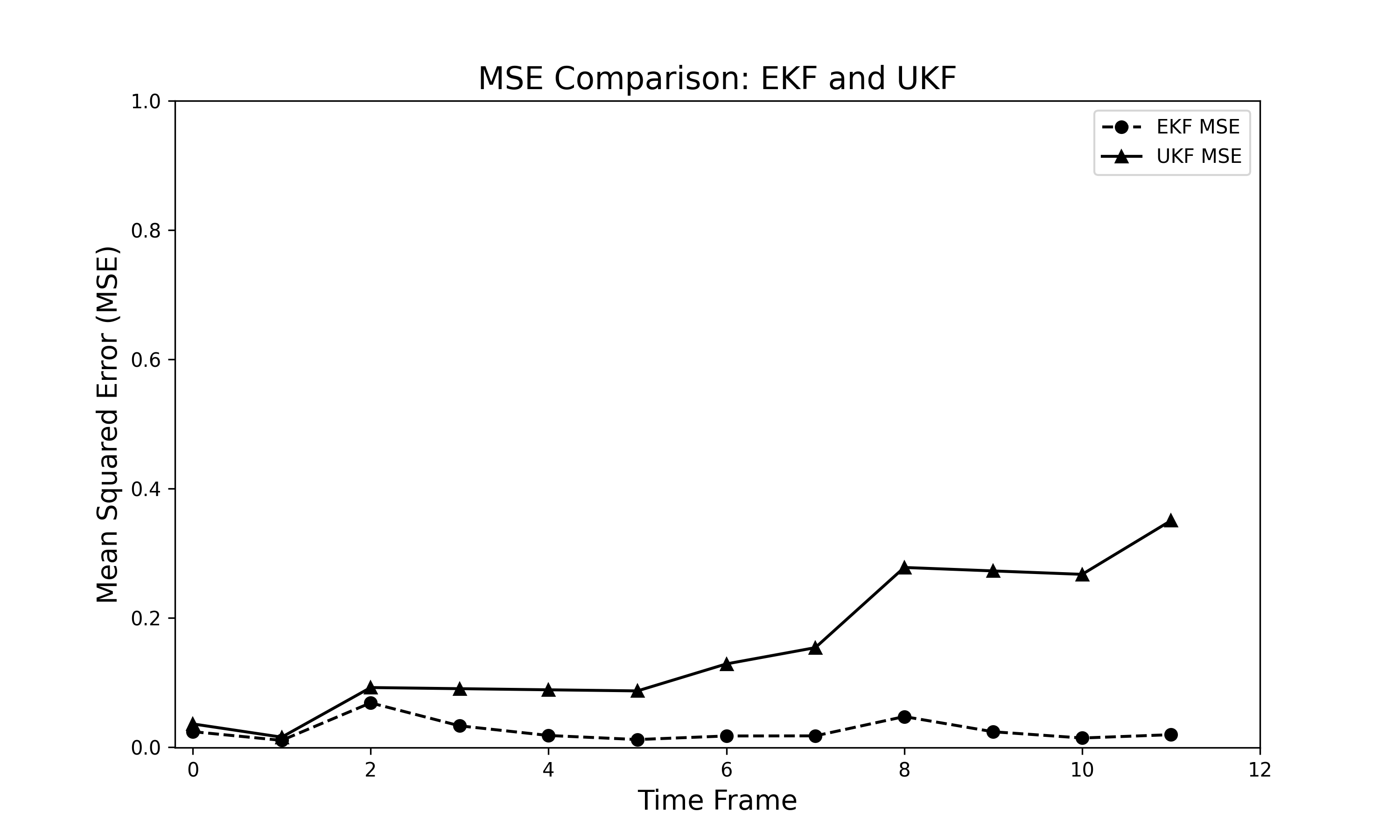} \\
		\includegraphics[width=0.45\textwidth]{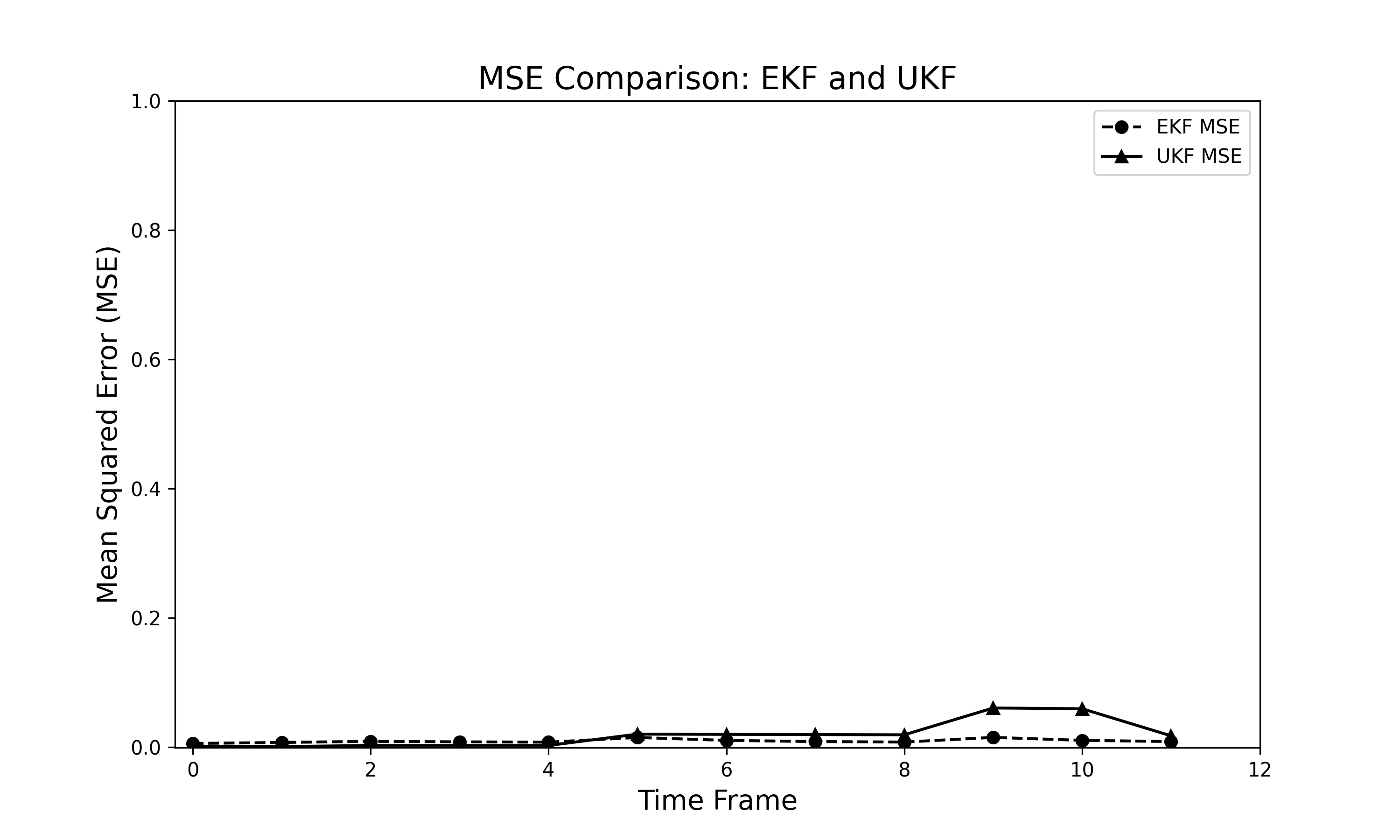} &\includegraphics[width=0.45\textwidth]{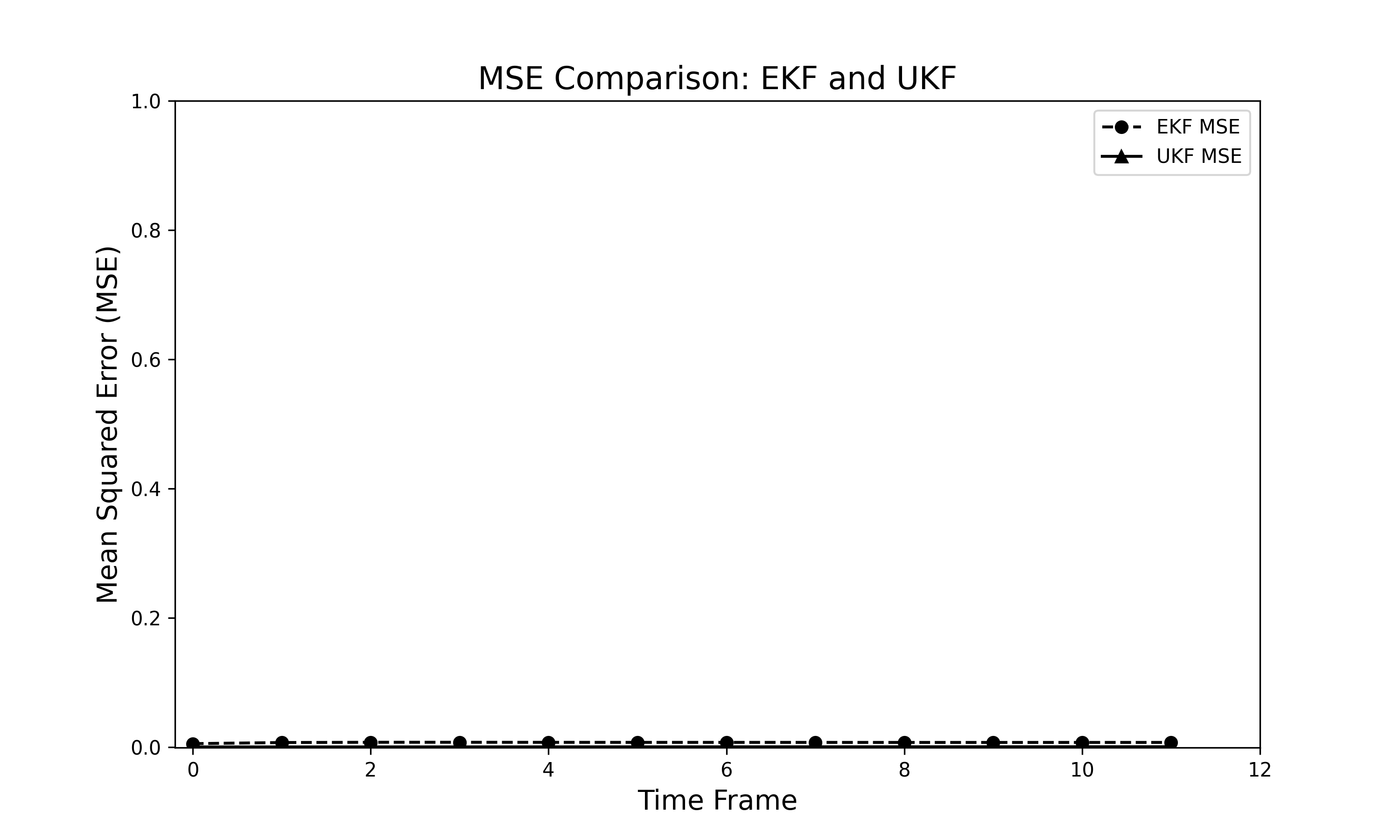}
	\end{tabular}
	\caption{[Color online] Comparison of MSE between the EKF and UKF in a stochastic setting, averaged across multiple realizations. The plot illustrates the MSE values for 3D facial point tracking over 12 time-framesfor users 1,2,3,4 (panels from the left to the right and from the top to the bottom). The EKF generally provides more accurate estimations in the presence of process noise and nonlinearities, as compared to UKF. The dataset used in this simulation is provided in \cite{Ariz2016novel}, licensed under a Creative Commons Attribution-NonCommercial-ShareAlike 4.0 International License.}
	\label{fig:8}
\end{figure}
\begin{figure}[h!]
	\centering
	\begin{tabular}{ll}
		\includegraphics[width=0.45\textwidth]{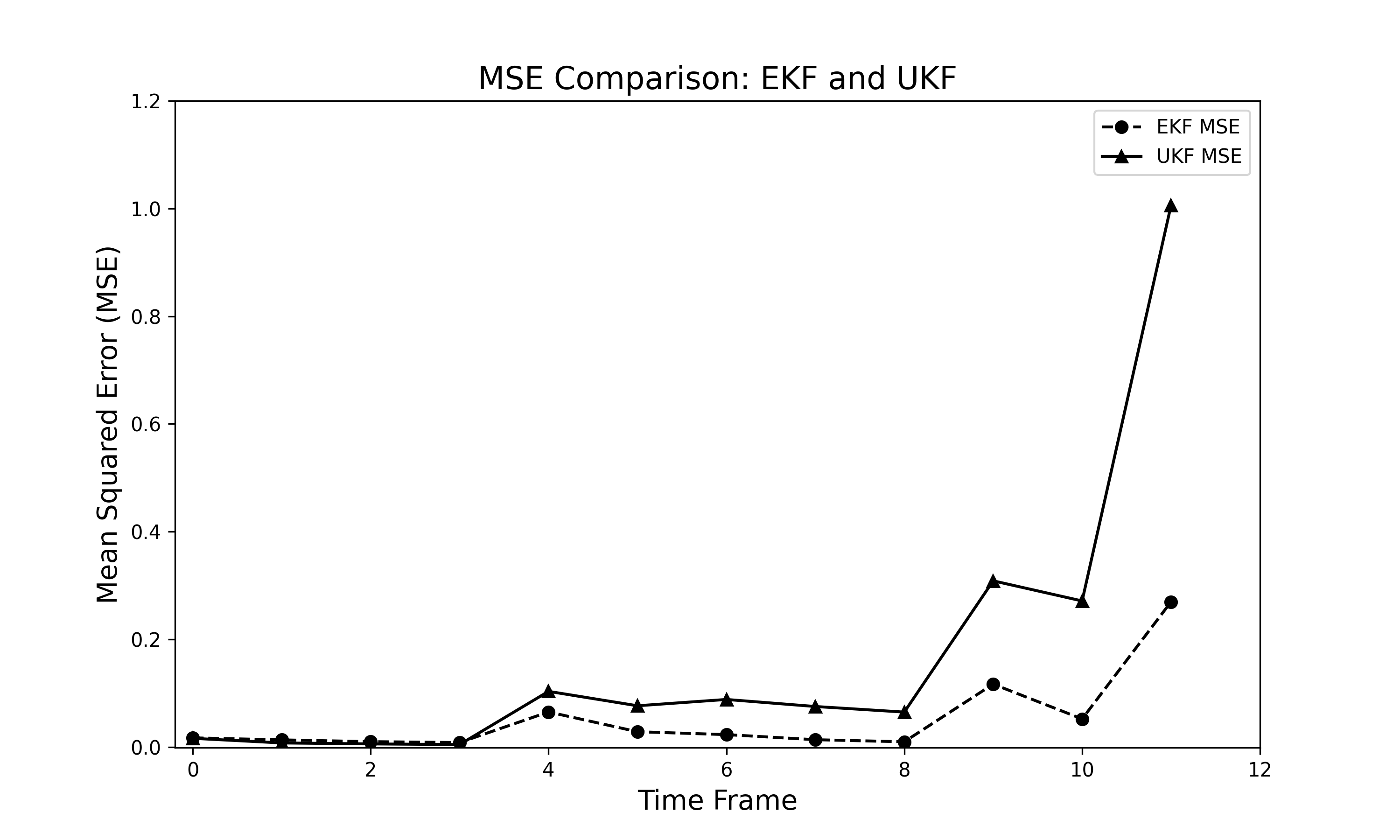} &\includegraphics[width=0.45\textwidth]{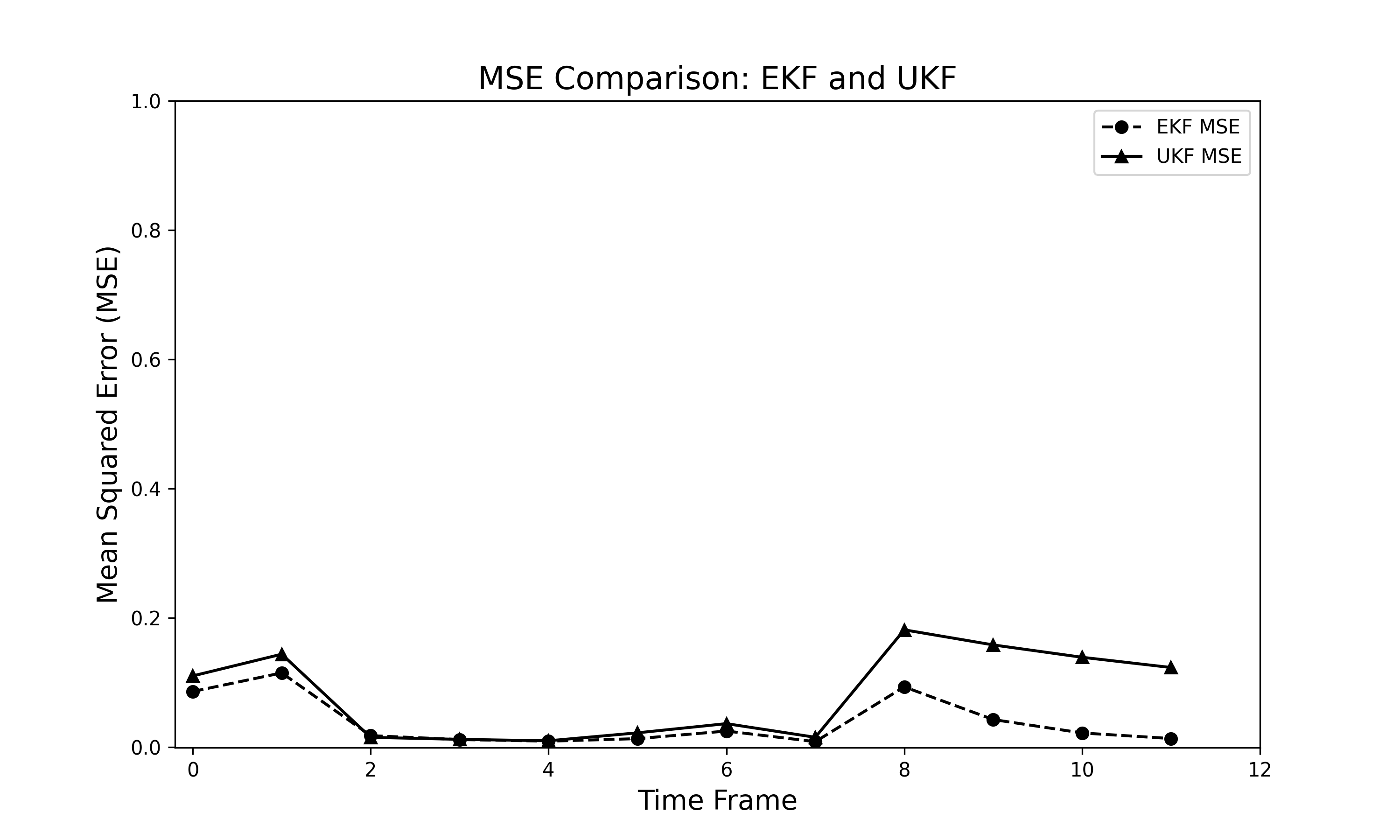} \\
		\includegraphics[width=0.45\textwidth]{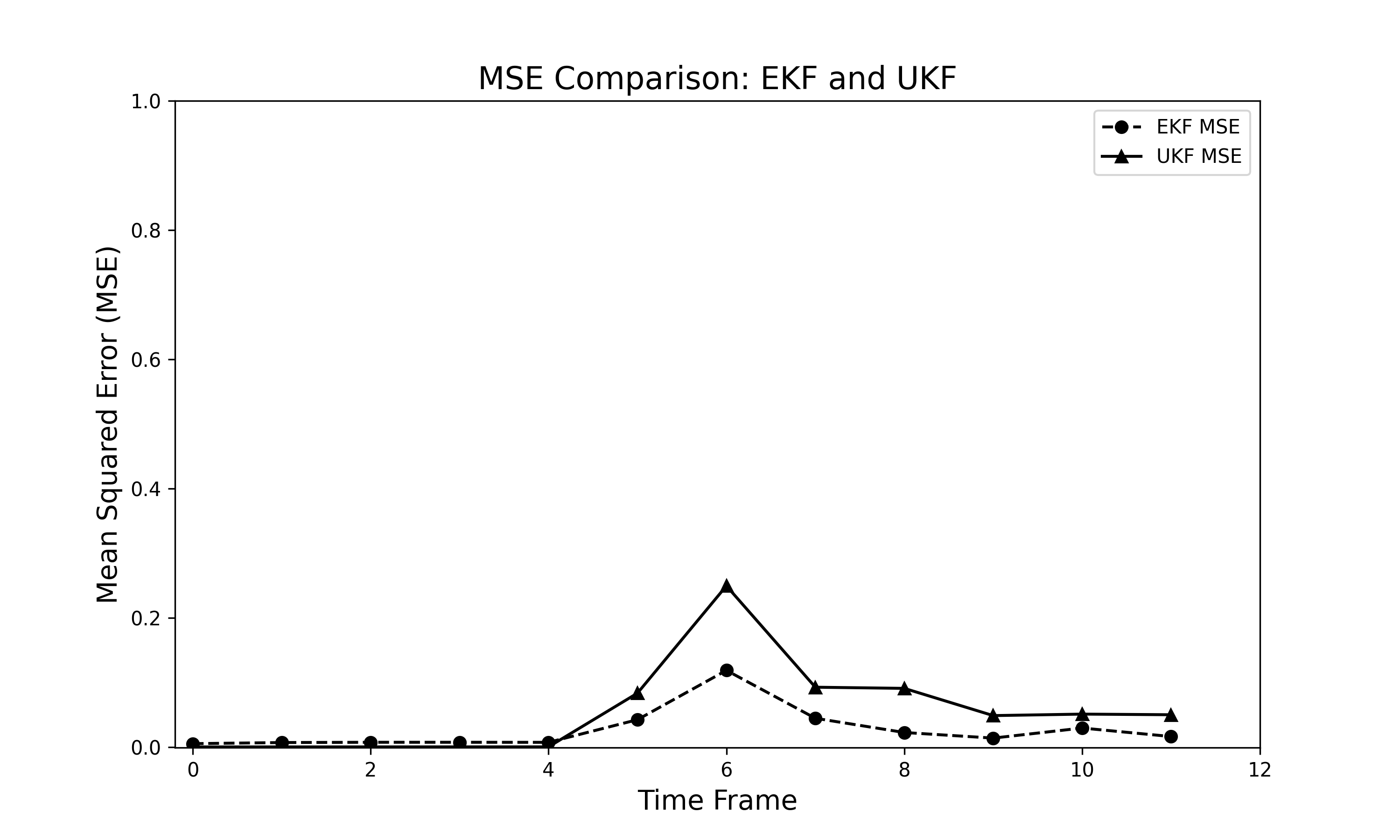} &\includegraphics[width=0.45\textwidth]{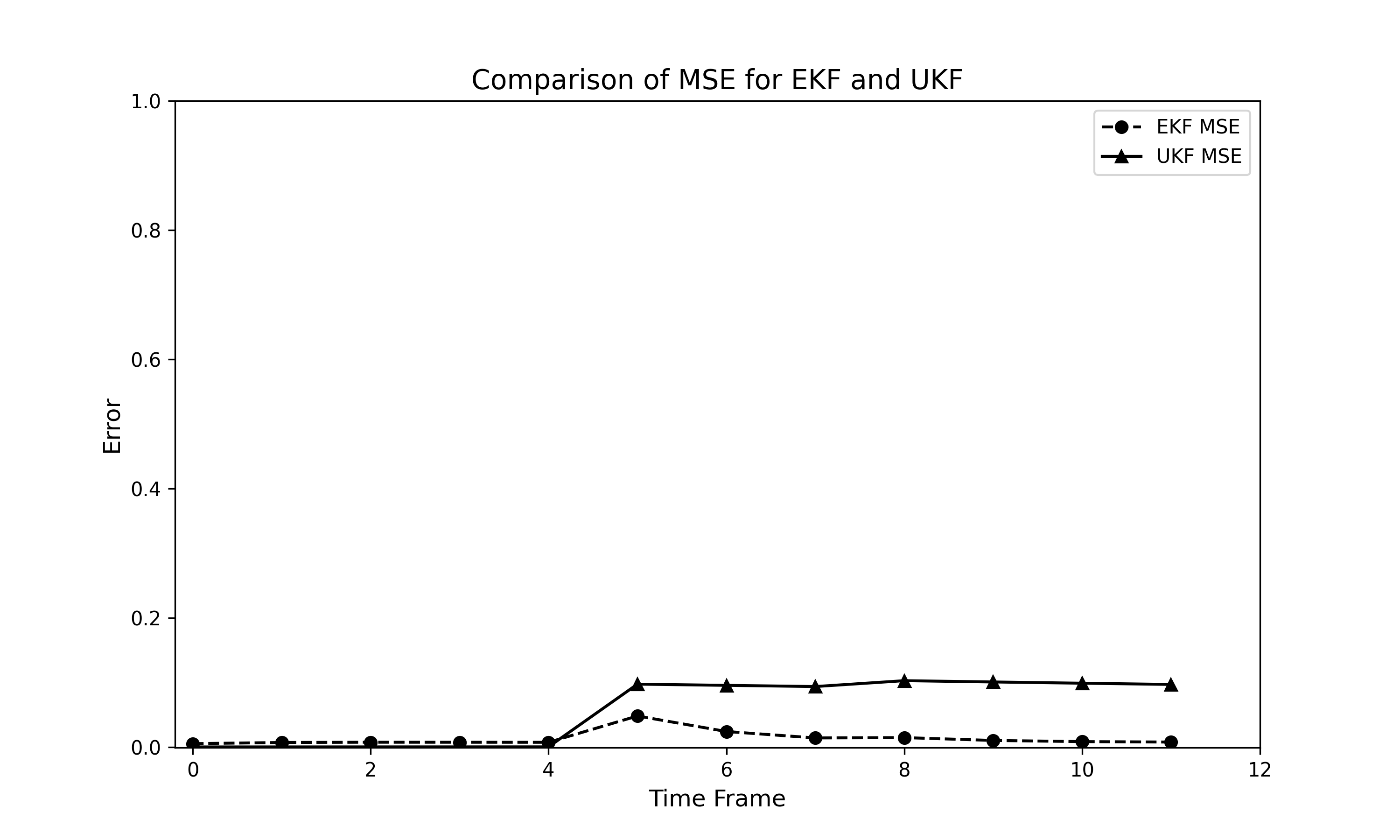}
	\end{tabular}
	\caption{[Color online] Comparison of MSE between the EKF and UKF in a stochastic setting, averaged across multiple realizations. The plot illustrates the MSE values for 3D facial point tracking over 12 time-framesfor users 5,6,7,8 (panels from the left to the right and from the top to the bottom). The EKF generally provides more accurate estimations in the presence of process noise and nonlinearities, as compared to UKF. The dataset used in this simulation is provided in \cite{Ariz2016novel}, licensed under a Creative Commons Attribution-NonCommercial-ShareAlike 4.0 International License. }
	\label{fig:9}
\end{figure}

\begin{figure}[h!]
	\centering
	\begin{tabular}{ll}
		\includegraphics[width=0.45\textwidth]{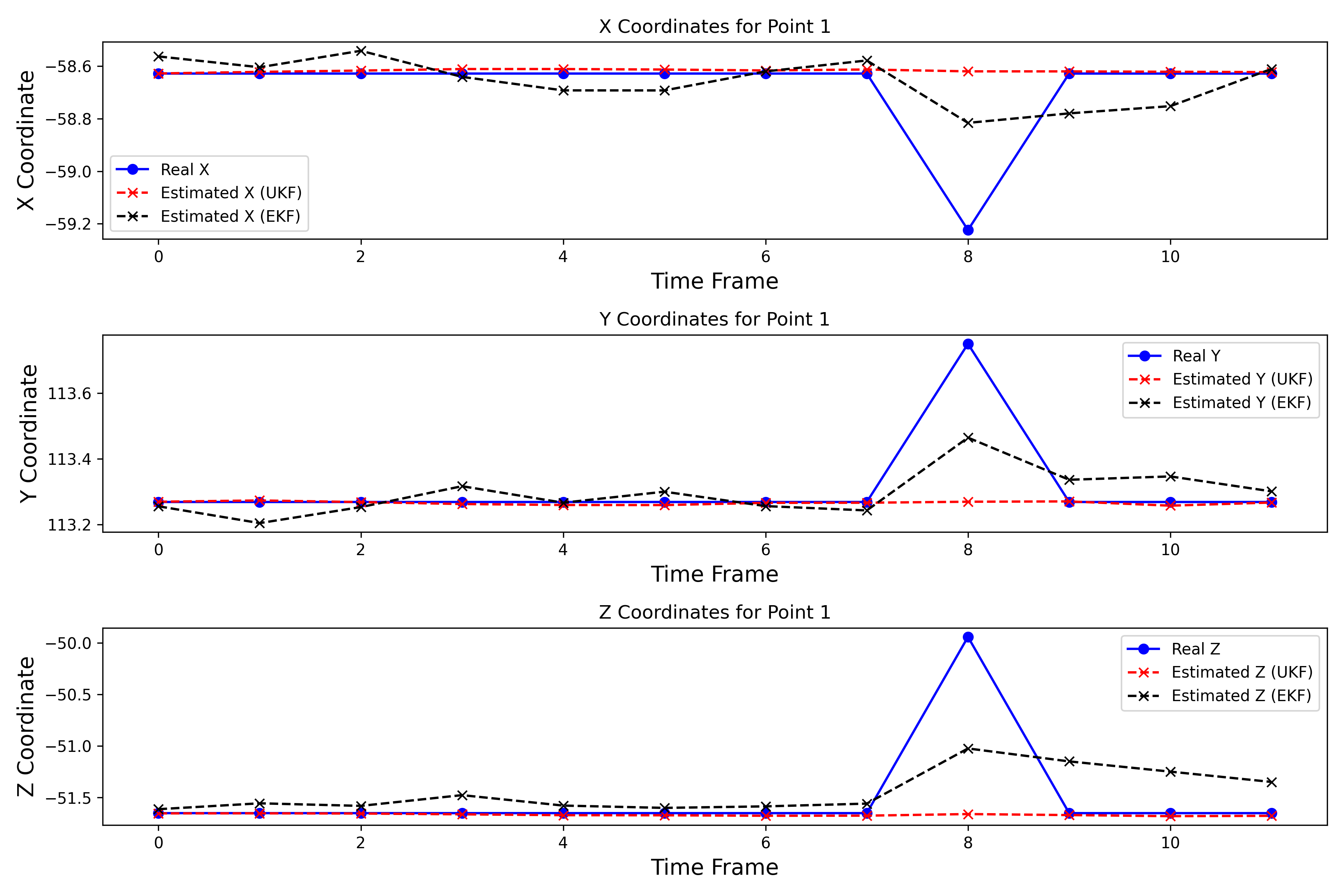} &\includegraphics[width=0.45\textwidth]{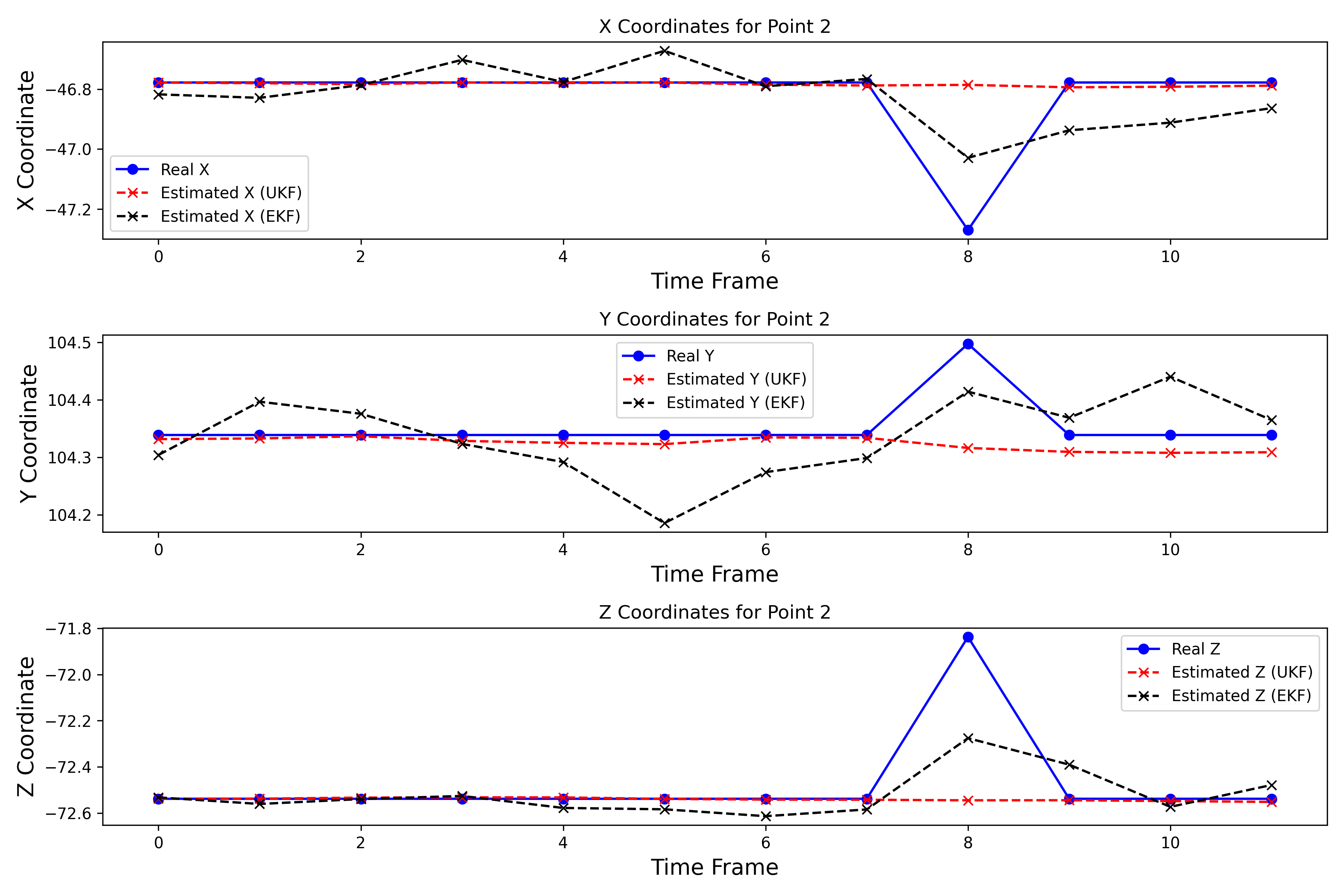} \\
		\includegraphics[width=0.45\textwidth]{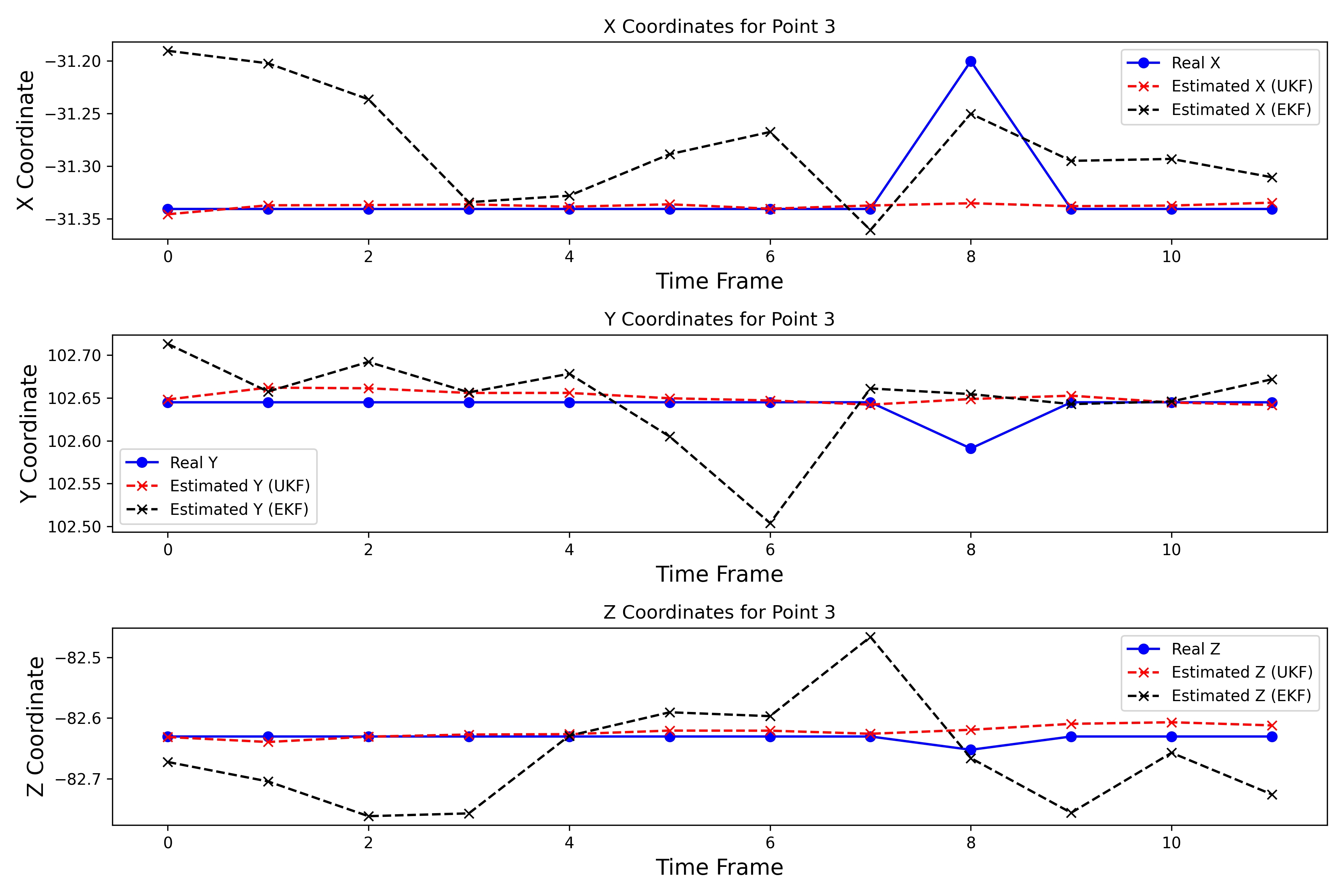} &\includegraphics[width=0.45\textwidth]{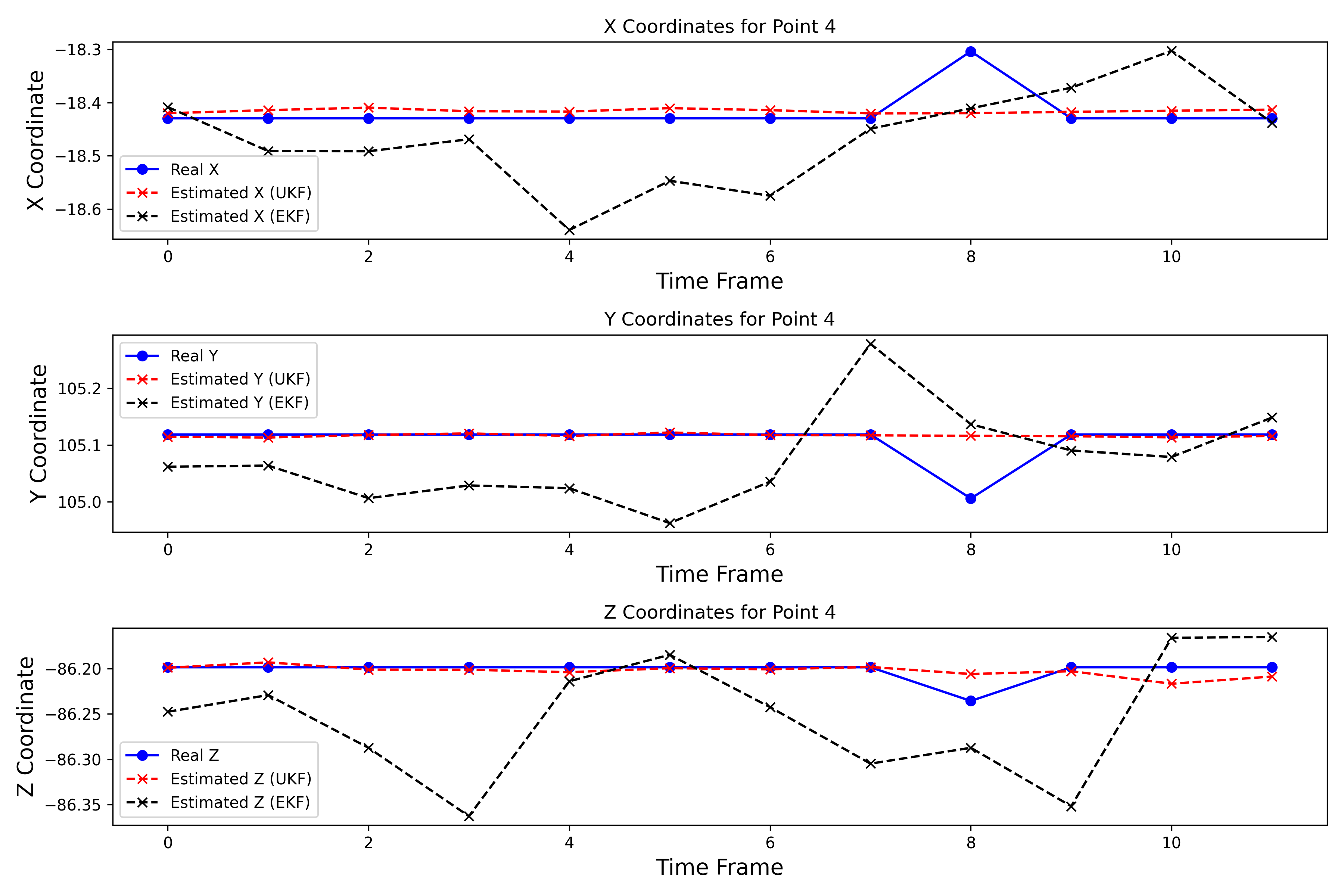}
	\end{tabular}
	\caption{[Color online] Comparison of real and estimated 3D facial landmark coordinates $x,y,z$ over the time frames for 4 selected points for user 1. The real data is plotted alongside the estimated values obtained using the stochastics UKF and EKF. Each subplot corresponds to a specific coordinate ($x, y$ or $z$), showing temporal variations across frames. The dataset used in this simulation is provided in  \cite{Ariz2016novel}, licensed under a Creative Commons Attribution-NonCommercial-ShareAlike 4.0 International License. }
	\label{fig:6}
\end{figure}

\begin{figure}[h!]
	\centering
	\begin{tabular}{ll}
		\includegraphics[width=0.45\textwidth]{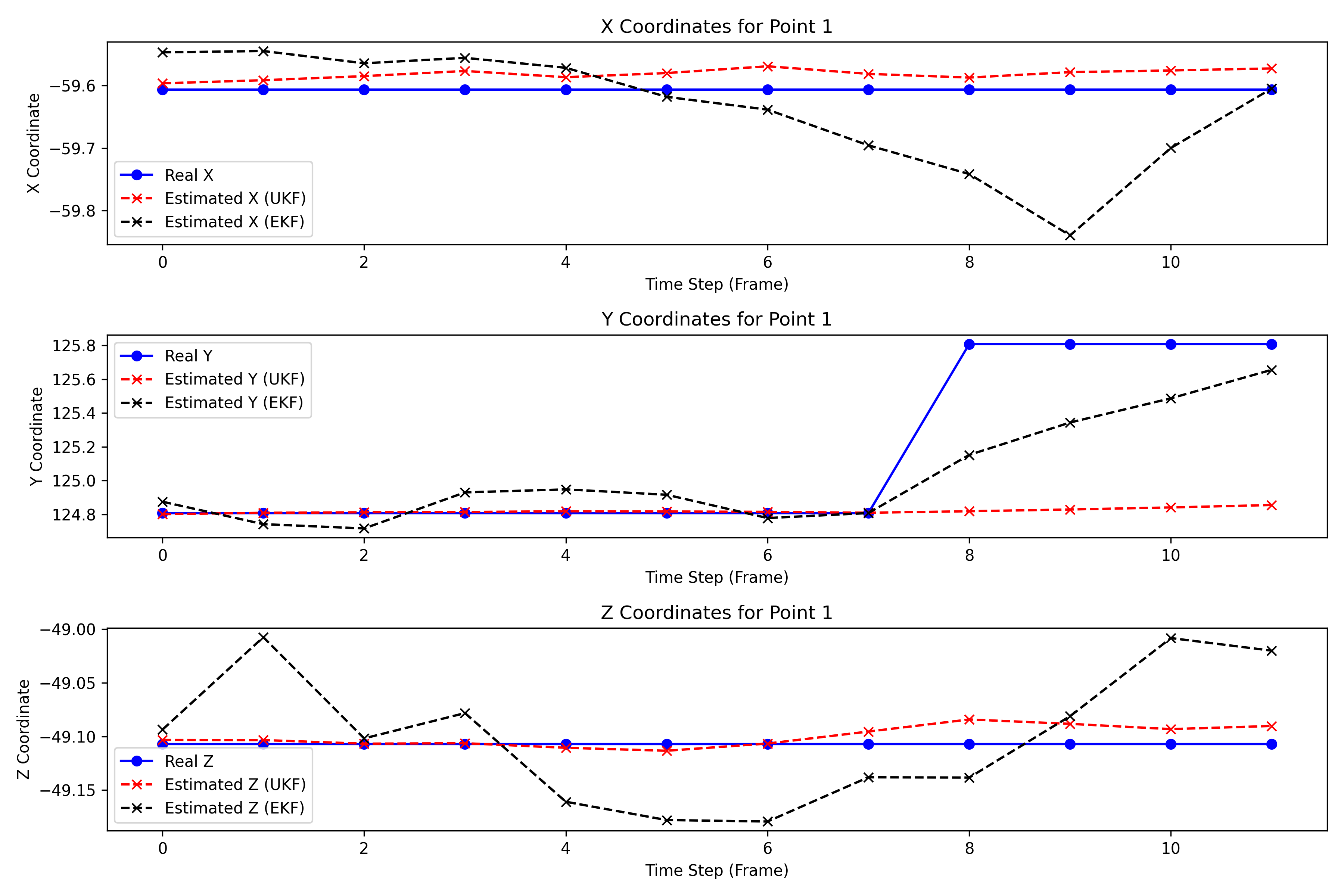} &\includegraphics[width=0.45\textwidth]{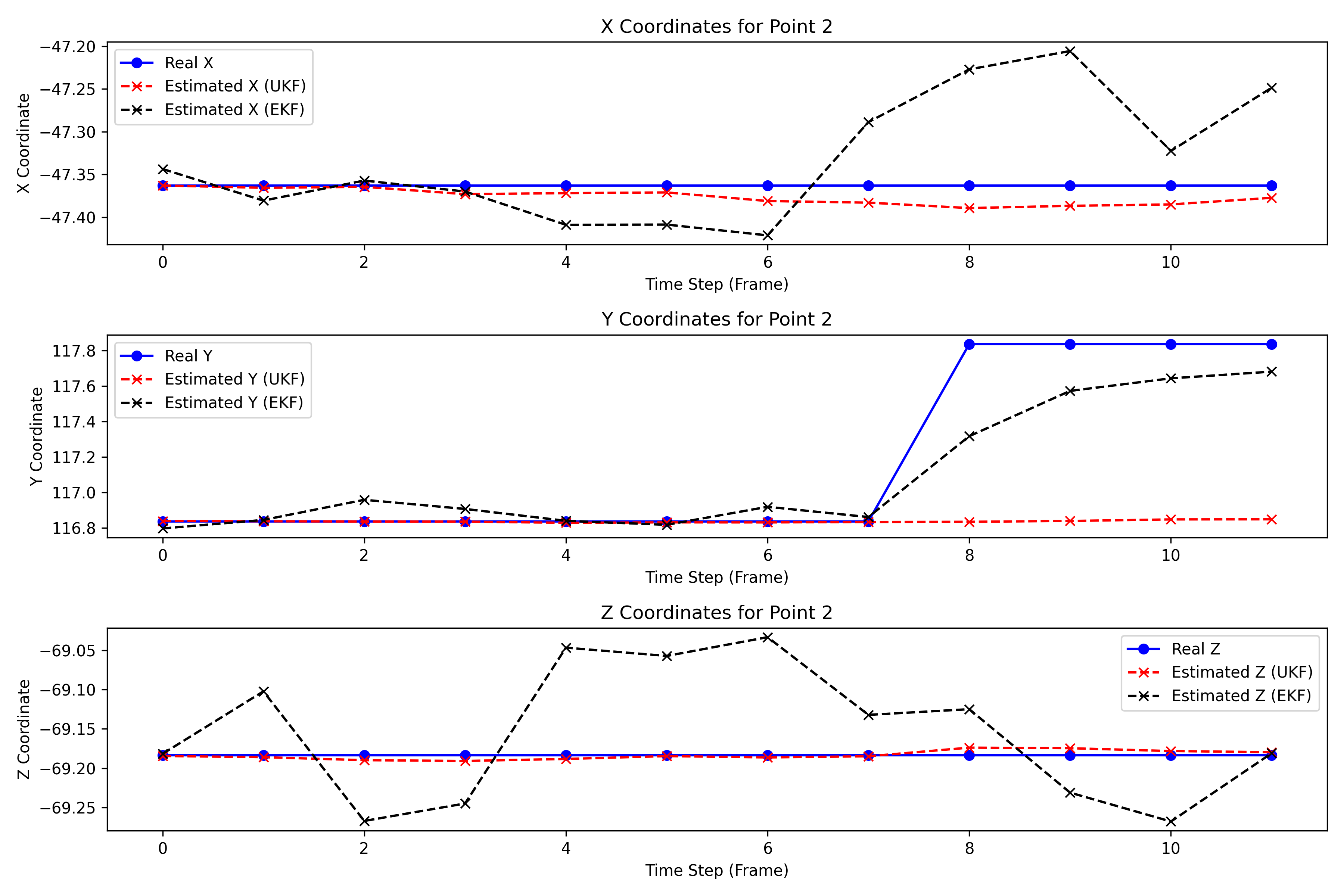} \\
		\includegraphics[width=0.45\textwidth]{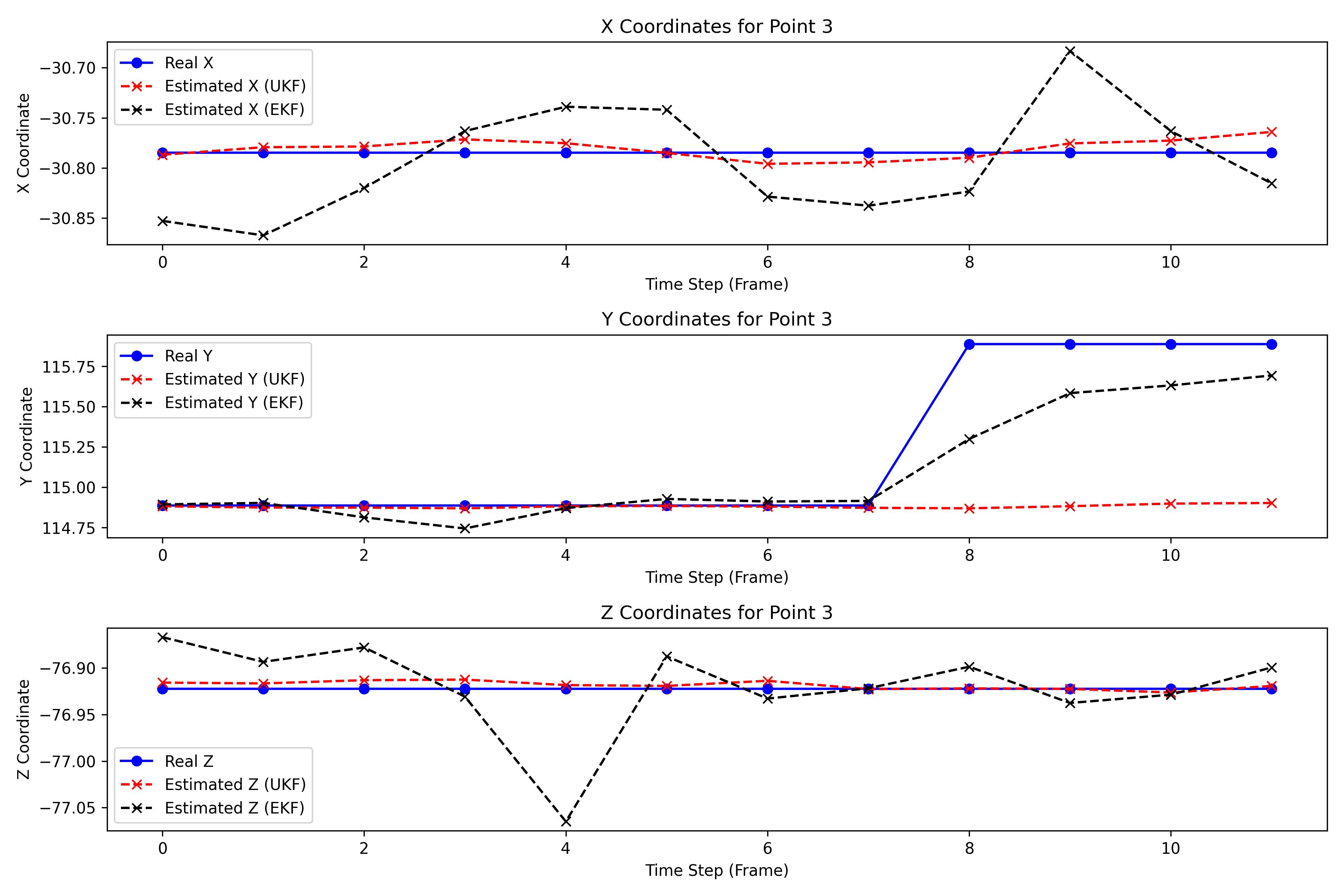} &\includegraphics[width=0.45\textwidth]{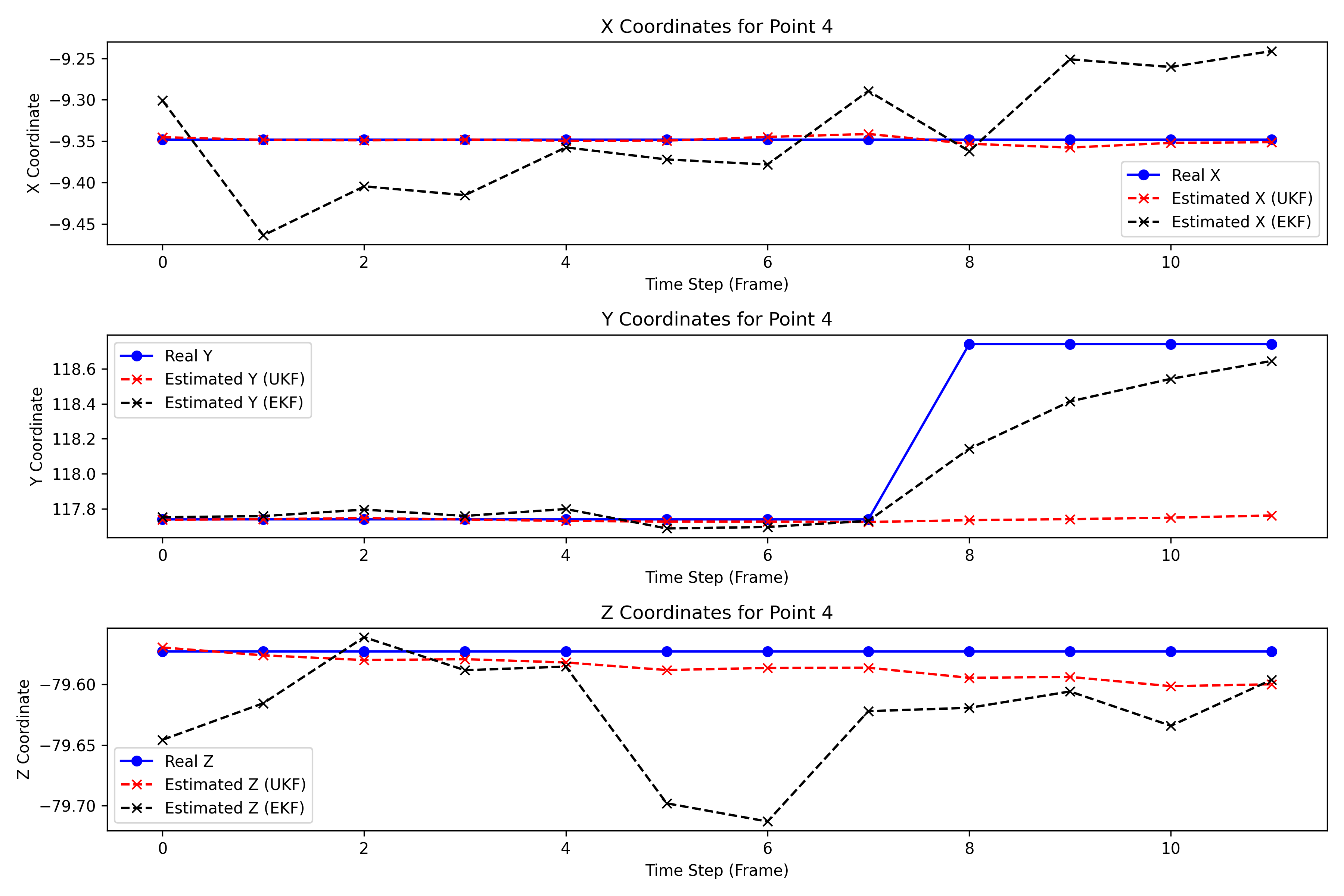}
	\end{tabular}
	\caption{[Color online] Comparison of real and estimated 3D facial landmark coordinates $x,y,z$ over the time frames for 4 selected points for user 2. The real data is plotted alongside the estimated values obtained using the stochastics UKF and EKF. Each subplot corresponds to a specific coordinate ($x, y$ or $z$), showing temporal variations across frames. The dataset used in this simulation is provided in  \cite{Ariz2016novel}, licensed under a Creative Commons Attribution-NonCommercial-ShareAlike 4.0 International License. }
	\label{fig:7}
\end{figure}

  In Figs. \ref{fig:8}-\ref{fig:9}, we observe that the EKF achieved a lower overall MSE compared to the UKF, with both filters maintaining MSE values close to zero in most cases. However, there are two instances where the UKF MSE is significantly higher. Specifically, in the top right panel of Fig. \ref{fig:8}, the UKF MSE reaches approximately 0.4 at the final time frame. Additionally, in the top right panel of Fig. \ref{fig:9}, the UKF MSE rises to around 0.4 in the second-to-last time frame and further increases to approximately 1 at the last time frame. In a noisy environment, we observed a role reversal as follows: EKF achieved lower overall MSE than UKF and UKF exhibited a large error (MSE $\sim1$) at a certain time frame. Moreover, in Figs. \ref{fig:6}-\ref{fig:7}, the EKF estimates demonstrate a closer fit to the real data compared to the UKF approximations across the time frames for four selected points from users 1 and 2. This observation is consistent with the results presented in Figs. \ref{fig:8}-\ref{fig:9}.


Why does EKF perform better in noisy conditions? UKF is more sensitive to noise. This is due to the fact that UKF propagates sigma points through the nonlinear model, it can amplify the effects of sensor noise, leading to tracking inaccuracies. If noise is heavy-tailed or non-Gaussian (e.g., caused by occlusions, shadows, or sudden motion), some sigma points may diverge, increasing error.
EKF’s reliance on the Jacobian matrix helps it filter out extreme noise effects more effectively than UKF. Why does UKF experience large errors in some frames?
If a subject moves suddenly, the process noise model may not accurately capture these changes.
%Temporary blocking of facial features (e.g., by hands, glasses, or shadows) can mislead UKF, causing abrupt spikes in error.
Anomalies in 3D depth sensor readings may disproportionately affect UKF’s sigma point predictions.
In real-world video footage or low-quality sensor data, where noise and occlusions are frequent, EKF provides more stable performance. UKF’s sensitivity to noise requires careful tuning of Q and R. In noisy environments, EKF may outperform UKF if the system is only mildly nonlinear. While UKF is more suited for handling strong nonlinearities due to its unscented transform, EKF’s linearization approach can still provide effective estimates when the system’s nonlinearity is less pronounced. Additionally, UKF can be more sensitive to noise, leading to higher MSE in some cases. This indicates that the degree of nonlinearity in the system significantly influences the performance of each filter, with EKF offering more robustness in noisy conditions for less nonlinear systems.

Based on our findings, the choice between EKF and UKF should be guided by the dataset characteristics and noise levels. UKF is preferable for high-accuracy applications with minimal noise, such as facial motion capture in controlled environments, while EKF offers greater robustness in unpredictable conditions, such as real-world video analysis with sensor noise. To enhance UKF performance, tuning noise parameters (Q, R) or implementing pre-filtering techniques like RANSAC to remove outliers can be beneficial. Additionally, a hybrid approach could be considered, where EKF is applied to noisy frames while UKF is used for smoother ones, optimizing accuracy and stability.

%Based on our findings, the choice between EKF and UKF should depend on the nature of the dataset and noise levels:
%
%Use UKF for high-accuracy applications with minimal noise, such as facial motion capture in controlled environments.
%
%Use EKF for robustness in unpredictable conditions, such as real-world video analysis with sensor noise.
%
%Improve UKF performance by tuning noise parameters (Q, R) or implementing pre-filtering techniques like RANSAC to remove outliers.
%
%Consider a hybrid approach, where EKF is used for noisy frames and UKF for smoother ones.

	\section{Conclusions and discussion}

In this study, we analyzed the performance of the EKF and UKF for tracking 3D facial landmarks using a dataset from \cite{Ariz2016novel}. Our results demonstrate that each filter's effectiveness depends on the dataset's characteristics and the noise conditions.

For deterministic cases in a noise-free environment, UKF consistently outperforms EKF by achieving lower mean squared error (MSE), demonstrating its superior accuracy in capturing nonlinear facial dynamics. However, we observed specific cases where UKF exhibited fluctuations in MSE, likely due to sensitivity to abrupt changes in the state transition, which EKF handled more robustly. In contrast, under noisier conditions, EKF demonstrated lower overall MSE, suggesting that it is better suited for handling unpredictable variations in real-world scenarios.

These findings highlight that the choice between EKF and UKF should be application-dependent. UKF is recommended for scenarios requiring high precision with minimal noise interference, such as controlled motion capture environments. On the other hand, EKF proves more reliable in practical settings where measurement noise is significant, such as real-time video-based facial tracking. To further optimize UKF performance, fine-tuning noise covariance parameters (Q, R) and employing pre-processing techniques like RANSAC for outlier removal could be beneficial. A potential hybrid approach, where EKF is applied to noisy frames while UKF is utilized for smoother segments, could leverage the strengths of both filters for improved robustness and accuracy.

These insights contribute to a better understanding of Kalman filtering techniques in 3D facial tracking and provide practical guidelines for selecting the appropriate filter based on environmental conditions and application requirements. Future research could explore adaptive filtering techniques that dynamically switch between EKF and UKF based on noise characteristics to improve tracking performance further. Additionally, incorporating deep learning-based denoising techniques before applying these filters could further enhance tracking performance in complex environments.

 \section*{Data availability statement}
 The open-access dataset UPNA is used in our study. \\The link is as follows: \href{https://www.unavarra.es/gi4e/databases/hpdb?languageId=1}{https://www.unavarra.es/gi4e/databases/hpdb?languageId=1}
 \section*{Conflict of interest}
 The authors declare that they have no conflict of interest.

% BibTeX users please use one of
%\bibliographystyle{spbasic}      % basic style, author-year citations
%\bibliographystyle{spmpsci}      % mathematics and physical sciences
%\bibliographystyle{spphys}       % APS-like style for physics
\bibliographystyle{splncs04}
\bibliography{mybibn-old}   % name your BibTeX data base

\end{document}